\documentclass[10pt,twocolumn,letterpaper]{article}

\usepackage[pagenumbers]{cvpr} 

\definecolor{cvprblue}{rgb}{0.21,0.49,0.74}
\usepackage[pagebackref,breaklinks,colorlinks,allcolors=cvprblue]{hyperref}
\newcommand{\myparagraph}[1]{\vspace{1mm}\noindent{\bf #1}}
\newcommand{\ourmethod}{\textit{CASS}}

\usepackage{array,multirow,graphicx}
\usepackage{float}
\usepackage{amsmath}
\usepackage{makecell}  
\usepackage{mathtools}
\usepackage{bbm}
\usepackage{hhline}
\usepackage{boldline}
\usepackage{algorithm} 
\usepackage{algpseudocode} 
\usepackage[ruled,vlined,algo2e]{algorithm2e}
\usepackage{listings}
\usepackage{graphicx}
\usepackage{amssymb}
\usepackage[dvipsnames]{xcolor}
\usepackage{colortbl}
\usepackage{booktabs} 
\usepackage[accsupp]{axessibility} 
\definecolor{customblue}{rgb}{0.294, 0.294, 0.882}
\definecolor{maroon}{cmyk}{0,0.87,0.68,0.32}

\usepackage{amssymb}
\usepackage{pifont}
\usepackage[ruled,vlined,algo2e]{algorithm2e}
\usepackage{listings}

\def\paperID{****} 
\def\confName{CVPR}
\def\confYear{2025}

\title{Distilling Spectral Graph for Object-Context Aware\\Open-Vocabulary Semantic Segmentation}

\author{
Chanyoung Kim$^{1}$\quad Dayun Ju$^{1}$\quad Woojung Han$^{1}$ \quad Ming-Hsuan Yang$^{1,2}$\quad Seong Jae Hwang$^{1}$\\ \\
$^{1}$Yonsei University\quad $^{2}$University of California, Merced
\\{\tt\small \{chanyoung, juda0707, dnwjddl, seongjae\}@yonsei.ac.kr},
{\tt\small mhyang@ucmerced.edu}
}

\begin{document}
\maketitle 
\begin{abstract}
Open-Vocabulary Semantic Segmentation (OVSS) has advanced with recent vision-language models (VLMs), enabling segmentation beyond predefined categories through various learning schemes.
Notably, training-free methods offer scalable, easily deployable solutions for handling unseen data, a key goal of OVSS.
Yet, a critical issue persists: lack of object-level context consideration when segmenting complex objects in the challenging environment of OVSS based on arbitrary query prompts.
This oversight limits models' ability to group semantically consistent elements within object and map them precisely to user-defined arbitrary classes.
In this work, we introduce a novel approach that overcomes this limitation by incorporating object-level contextual knowledge within images. 
Specifically, our model enhances intra-object consistency by distilling spectral-driven features from vision foundation models into the attention mechanism of the visual encoder, enabling semantically coherent components to form a single object mask.
Additionally, we refine the text embeddings with zero-shot object presence likelihood to ensure accurate alignment with the specific objects represented in the images.
By leveraging object-level contextual knowledge, our proposed approach achieves state-of-the-art performance with strong generalizability across diverse datasets.
\let\thefootnote\relax\footnote{\scriptsize{Project Page:~\url{https://micv-yonsei.github.io/cass/}}}

\end{abstract}    
\begin{figure}[t!]
  \centering
   \includegraphics[width=\linewidth]{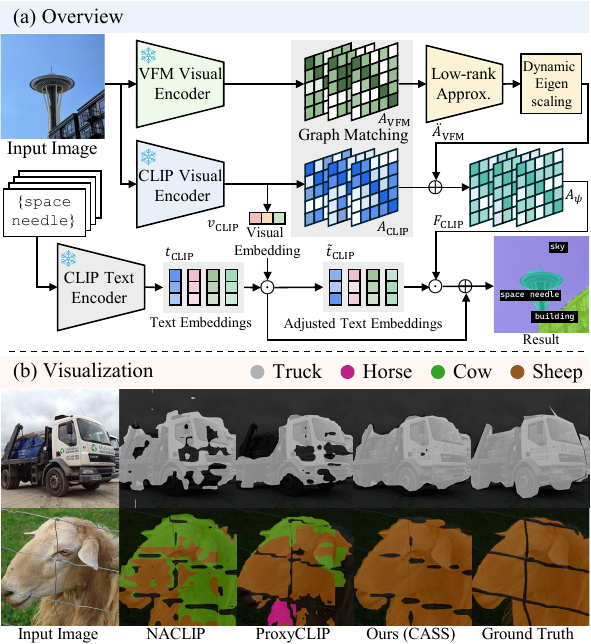}
    \vspace{-18 pt}
   \caption{
We present {\ourmethod}, object-level Context-Aware \textit{training-free} open-vocabulary Semantic Segmentation model. (a) \textit{Overview}:
Our method distills the vision foundation model's (VFM) object-level contextual spectral graph into CLIP’s attention and refines query text embeddings towards object-specific semantics.
(b) \textit{Object-Level Context}:
This result illustrates how incorporating object-level context improves segmentation accuracy by unifying object-wise components into a user-defined object class; for example, (top row) our approach precisely segments the truck’s body, wheels, and cargo area and (bottom row) accurately groups elements into a single object, such as the sheep, whereas baselines~\cite{hajimiri2025naclip,lan2024proxyclip} often fail to achieve this unified segmentation.
}
   \label{fig:example}
\vspace{-10 pt}
\end{figure}

\vspace{-10pt}
\section{Introduction}
\label{sec:intro}
Open-vocabulary semantic segmentation (OVSS)~\cite{liang2023open,xu2023open,cha2023learning,xu2022groupvit,zhou2022extract} aims to predict pixel-level labels for \textit{arbitrary prompts} defined by user input (e.g., the proper noun \texttt{Space Needle} in Fig.~\ref{fig:example}(a)), advancing beyond previous models scoped to predefined static classes~\cite{ronneberger2015u,hatamizadeh2022unetr,chen2018encoder,cobra2025}. 
To achieve this, fully supervised OVSS models~\cite{cho2024cat,xu2023side,xie2024sed} enhance generalization to seen classes using labeled training data.
Yet, their reliance on seen classes during training may risk overfitting and limit scalability, as labeled data is required for retraining to adapt to new domains~\cite{shan2024open}.

To better address this, recent studies~\cite{wang2023sclip, lan2024proxyclip, hajimiri2025naclip, zhou2022extract,lan2024clearclip} propose \textit{training-free} OVSS methods that leverage pretrained vision-language models (VLMs) (e.g., CLIP~\cite{radford2021learning}) to integrate aligned visual and textual representations.
These works focus on improving semantic association between visual patches in CLIP image encoder to produce more precise segmentation maps, achieving highly promising results.
This {training-free} scheme offers key advantages: it (1) generalizes well to unseen classes without requiring additional labeled data, making it more scalable, thus (2) allows direct adaptation to dynamic real-world applications.
Capitalizing on these benefits, training-free OVSS aligns with OVSS’s core goal of adaptability across diverse domains, enabling it to reach its full potential.

We tackle OVSS in \textit{a training-free} manner, presenting a challenging and practical approach aligned with OVSS’s true objectives.
Although existing training-free methods~\cite{lan2024proxyclip,wang2023sclip,sun2024clip,kang2024defense} perform well on numerous tasks, they all share a common limitation: the lack of \textit{``object-level context."}
To understand \textit{``object-level context"}, consider a \texttt{truck} and \texttt{sheep} in Fig.~\ref{fig:example}(b).
Components comprising a \texttt{truck}, such as its wheels and cargo area, should be grouped as a single entity. 
Similarly, components of a \texttt{sheep} should be accurately unified under the correct user-given object class. 
However, existing training-free OVSS methods, which use CLIP as the image backbone for segmentation, often struggle to capture object-level context, failing to group object components into a single, user-defined object.
These challenges stem from CLIP’s focus on learning global image semantics, which may be insufficient for capturing dense object-level semantics~\cite{wang2023sclip,hajimiri2025naclip,zhou2022extract}.
Thus, to achieve accurate training-free OVSS, a proper object-level context must be considered in CLIP.

We address the challenges of training-free OVSS by introducing the object-level context and propose the Context-Aware Semantic Segmentation ({\ourmethod}) model. 
Vision foundation models (VFMs), such as DINO~\cite{caron2021emerging,oquab2023dinov2}, capture fine-grained, patch-level semantics but lack object-level context, making them unsuitable for direct use in tasks requiring such context.
Recent studies~\cite{eagle2024,melas2022deep} have shown that applying graph spectral techniques to VFM attention graphs can transform patch-level representations into object-level representations. 
Accordingly, we apply spectral techniques to extract essential object-level context from the VFM attention graph.
The extracted spectral features can be used directly for segmentation (e.g., clustering eigenvectors); however, this approach may not be suitable for OVSS, where textual alignment is essential.
Thus, we propose distilling low-rank components of VFM into CLIP to enhance its object-level context understanding, while maintaining alignment with textual information.
Specifically, we decompose the VFM’s attention graph to extract low-rank components, filtering irrelevant information while emphasizing essential object-level context.
These low-rank components are then distilled into CLIP’s attention to embed the fundamental object structure of the VFM graph.

Our model also leverages CLIP's highly effective zero-shot object classification capability (i.e., object presence prior), widely validated in prior work~\cite{radford2021learning,jiang2023clip,mao2023clip4hoi}, to capture detailed object-level context within scenes.
As such, we adjust the text embeddings based on the object presence prior encoded by CLIP.
This process involves refining the embeddings to better align with object-specific semantics.
Then, we refine patch-text similarities using the object presence prior, ensuring that the final segmentation map reflects an object-specific perspective.
Our {\ourmethod}, carefully designed to enhance object-level contextual knowledge, captures richer intra-object coherence and enables cohesive groupings of semantically related elements.

The main contributions of this work are:
\begin{itemize}
\item{We enhance object-level context by distilling rich semantic understanding from VFM into the CLIP attention mechanism, facilitating a more consistent grouping of each object into a unified semantic entity.}
\item{We present object presence prior-driven text embeddings and patch-text similarity refinement, promoting the model to precisely classify objects.}
\item{We demonstrate the performance of our model in various semantic segmentation datasets and achieve state-of-the-art performance in training-free OVSS.}

\end{itemize}
\vspace{-5pt}
\section{Related Works}
\label{sec:related_work}

\begin{figure*}[t!]
    \begin{center}
        \includegraphics[width=\textwidth]{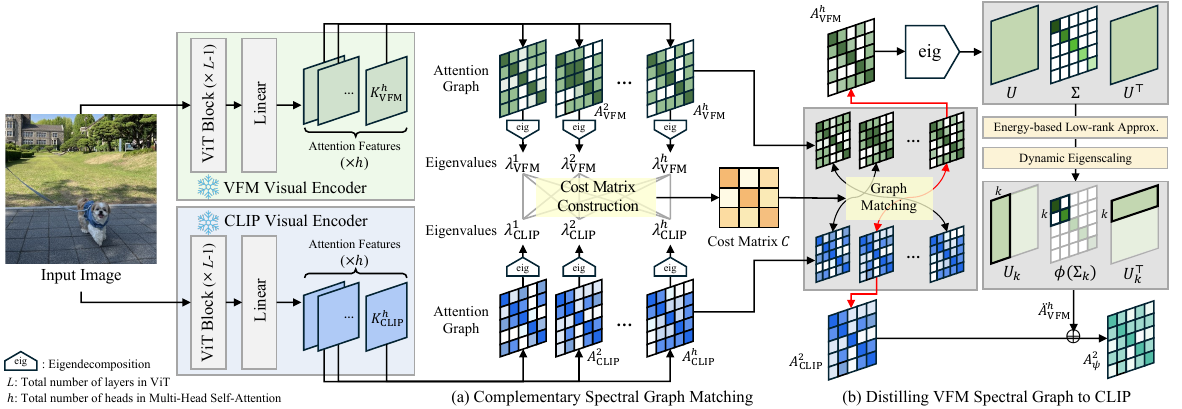}
    \end{center}
    \vspace{-17pt}
    \caption{
    Detailed illustration of our proposed \textit{training-free} spectral object-level context distillation mechanism in Sec.~\ref{sec:method_local}. 
    By matching the attention graphs of VFM and CLIP head-by-head to establish complementary relationships, and distilling the fundamental object-level context of the VFM graph to CLIP, we enhance CLIP’s ability to capture intra-object contextual coherence.
    }
    \label{fig:local}
    \vspace{-10pt}
\end{figure*}

\vspace{-3pt}
\paragraph{Open-Vocabulary Semantic Segmentation.}

Open-vocabulary semantic segmentation (OVSS) aims to segment images corresponding to arbitrary user-given prompts, using vision-language aligned representations (e.g., CLIP~\cite{radford2021learning}). 
Meanwhile, training-free approaches~\cite{Barsellotti_2024_WACV,barsellotti2024training,lan2024proxyclip,zhou2022extract,kang2025your} have risen for their ability to adapt directly to new tasks without extra training.
Among these, methods~\cite{wang2023sclip,zhou2022extract,lan2024proxyclip,hajimiri2025naclip} that do not rely on additional datasets have gained attention, as they require no extra adaptation effort, aligning better with the objective of OVSS.
Earlier approaches investigated leveraging CLIP features for better localization~\cite{zhou2022extract,li2023clip} or for grounding proposed masks~\cite{sun2024clip,kang2024defense}.
SCLIP~\cite{wang2023sclip} uses correlative self-attention to improve inter-patch correlations. 
NACLIP~\cite{hajimiri2025naclip} applies a Gaussian kernel to attention maps for better spatial consistency, and ProxyCLIP~\cite{lan2024proxyclip} integrates patch-wise consistent features from DINO~\cite{caron2021emerging}. 
Although these methods demonstrated strong results, they often struggle to group different parts of the same object due to a lack of explicit dense label knowledge.
To this end, we attribute the underlying challenges of training-free OVSS by enriching the object-level contextual understanding.

\myparagraph{Vision Foundation Models.}
Vision Foundation Models (VFMs), especially vision transformer (ViT)-based~\cite{dosovitskiy2021an} self-supervised models (e.g., DINO~\cite{caron2021emerging,oquab2023dinov2}), have significantly advanced computer vision through their ability to capture patch-wise semantic understanding at both the feature and attention levels.
Among various downstream tasks~\cite{hamilton2022unsupervised,seong2023leveraging,simeoni2021localizing,wang2023cut,mariotti2024improving} leveraging VFM's capability, one line of research focuses on leveraging the attention components in VFM~\cite{simeoni2021localizing,eagle2024,melas2022deep}, which exhibit a detailed understanding of patch-wise semantic interaction between different image regions.
Thus, while ProxyCLIP~\cite{lan2024proxyclip} utilizes VFM features, our method takes a different effort by treating the attention mechanism in VFM as a graph. 
We then extract core structural patterns of the VFM graph to emphasize object-level context, providing an effective solution for producing precise object masks in training-free OVSS.

\myparagraph{Object-Level Semantic Segmentation.}
Semantic segmentation groups object components that have similar semantics into a single mask, traditionally achieved through models trained on extensive human-labeled, pixel-wise annotations~\cite{hatamizadeh2022unetr,chen2018encoder,ronneberger2015u,kirillov2023segment}.
However, despite recent advancements in unsupervised semantic segmentation~\cite{hamilton2022unsupervised,seong2023leveraging} (USS) that do not rely on dense labels, a core challenge persists: individual object components are not consistently grouped into a single entity.
To address this issue, recent USS studies~\cite{eagle2024,wen2022self,seitzer2023bridging} have incorporated object-level representation learning to improve the model’s ability to merge single objects into unified entities.
However, these efforts have not been considered in the training-free OVSS task, which faces similar challenges in an unsupervised context.
Therefore, we focus on enriching object-level context in training-free OVSS to achieve a more effective grouping of components within each object, enabling clearer and more consistent segmentation across diverse visual domains.

\vspace{-5pt}
\section{Method}
\label{sec:method}

We detail our proposed model, {\ourmethod}, implemented in a training-free manner, as outlined in Fig~\ref{fig:example}(a). 
Our model jointly enhances object-level contextual understanding with two primary aspects: {(1)} {Spectral Object-Level Context Distillation} (Sec.~\ref{sec:method_local}) and {(2)} {Object Presence-Driven Object-Level Context} (Sec.~\ref{sec:method_global}). 
We first outline the basic pipeline of training-free OVSS in Sec.~\ref{sec:method_prelim}.

\subsection{Preliminaries}
\label{sec:method_prelim}

\vspace{-5pt}
\myparagraph{CLIP Visual Encoder.}
Training-free OVSS models~\cite{wang2023sclip, hajimiri2025naclip, lan2024proxyclip} leverage CLIP~\cite{radford2021learning} for its rich alignment of visual and linguistic information. 
However, the CLIP ViT visual encoder $\mathcal{F}_\text{CLIP}^v$ focuses on aligning the \texttt{[CLS]} token with text embeddings during training, often neglecting the patch-wise spatial interactions among visual tokens that are crucial for dense prediction~\cite{wang2023sclip, hajimiri2025naclip, shao2024explore}.
To produce CLIP feature representations more suitable for semantic segmentation tasks, we remove residual connection, feed-forward network, and self-attention layer in the final layer of $\mathcal{F}_\text{CLIP}^v$, following recent works~\cite{wang2023sclip, hajimiri2025naclip}.
Specifically, when the set of visual tokens ${Z}= \text{Concat}\left[z_1, \dots, z_N \right] \in \mathbb{R}^{N\times D_{Z}}$, where $N$ is the total number of patches except \texttt{[CLS]} token and $D_{Z}$ denotes patch dimension space, is input into the final block of the ViT, the following operations are as
\vspace{-5pt}
\begin{equation}
{Z}^* = \text{SA}(\text{LN}({Z})),
\label{eq:first}
\vspace{-5pt}
\end{equation}
where SA and LN represent Self-Attention and Layer Normalization, respectively.
The modified SA layer is
\vspace{-5pt}
\begin{equation}
{M}_\text{CLIP}^i = \text{softmax}\left({{A}_\text{CLIP}^i}/{\sqrt{D_{h}}}\right),
\label{eq:softmax}
\end{equation}

\vspace{-20pt}

\begin{equation}
{Z}^* = \text{Concat}\left[ {M}_\text{CLIP}^1 {V}_\text{CLIP}^1, \dots, {M}_\text{CLIP}^h {V}_\text{CLIP}^h \right] W^O,
\end{equation}
where $D_{h}$ denotes the head dimension, $h$ is the number of heads, and $W^O \in \mathbb{R}^{D_{Z} \times D_{Z}}$ is the output projection matrix.
We define CLIP attention map for $i$-th head as ${A}_{\text{CLIP}}^i = {K}_{\text{CLIP}}^i {K}_{\text{CLIP}}^{i\top} \in \mathbb{R}^{N \times N}$, where attention key ${K}_{\text{CLIP}} \in \mathbb{R}^{N \times D_h \times h}$ and value ${V}_{\text{CLIP}} \in \mathbb{R}^{N \times D_h \times h}$ result from the linear transformation of $\text{LN}({Z})$.

\begin{figure}[t!]
  \centering
   \includegraphics[width=\linewidth]{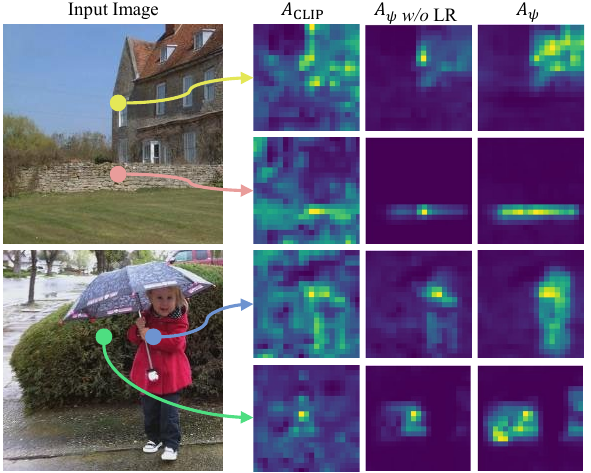}
   \vspace{-20pt}
   \caption{
   Attention score visualization for various query points. Left: Vanilla CLIP ($ {A}_{\text{CLIP}}$) shows noisy, unfocused attention. Center: VFM-to-CLIP distillation \textit{without} low-rank eigenscaling shows partial object grouping with limited detail. Right: Incorporating our low-rank eigenscaling (Eq.~\eqref{eq:distillation}) captures object-level context, improving grouping within a single object.
   }
   \label{fig:attn_vis}
   \vspace{-10pt}
\end{figure}

\myparagraph{Inference Pipeline.} 
We use a sliding window inference to ensure that ViT can capture detailed parts of the image by dividing the entire image $\mathcal{I}$ into smaller windows $\hat{\mathcal{I}}$. 
The visual tokens ${Z}^* \in \mathbb{R}^{N\times D_{Z}}$ are then computed from each $\hat{\mathcal{I}}$ through $\mathcal{F}_\text{CLIP}^v$ as in Eq.~\eqref{eq:first}. 
To perform OVSS with given texts, we project the visual tokens $ {Z}^*$ into the vision-language latent space as ${F}_\text{CLIP} = \mathcal{P}_{\text{CLIP}}({Z}^*) \in \mathbb{R}^{N \times d}$, where $\mathcal{P}_{\text{CLIP}}$ is a projection matrix into CLIP latent space and $d$ is the CLIP latent dimension.
We compute patch-text similarity $\hat{\mathcal{S}} \in \mathbb{R}^{N \times C}$ for each image window $\hat{\mathcal{I}}$ as
\vspace{-1pt}
\begin{equation}
    \hat{\mathcal{S}} = {F}_\text{CLIP} \left[ \left\{ t_\text{CLIP}^i \right\}_{i=1}^C{} \right]^{\top},
    \label{eq:logit}
    \vspace{-1pt}
\end{equation}
where ${\{ t_\text{CLIP}^i }\}_{i=1}^C := \text{Concat}\left[t_\text{CLIP}^1, \dots, t_\text{CLIP}^C \right] \in \mathbb{R}^{C \times d}$ denotes the encoded text embeddings, and $C$ is the number of classes.
Finally, the results of all windows are combined to create a segmentation map for the entire image.

\subsection{Spectral Object-Level Context Distillation}
\label{sec:method_local}
Our model {\ourmethod} aims to capture such object-level contextual interaction within the image features to associate different parts of the same object.
However, the limited ability of CLIP to capture semantic relationships between patches often neglects the object-level context, as illustrated in the left column of Fig.~\ref{fig:attn_vis} (${A}_\text{CLIP}$).
To address this limitation, we leverage VFM (e.g., DINO~\cite{caron2021emerging}), enabling a deeper object-level contextual understanding.
Among VFM features, we utilize VFM attention key component ${K}_{\text{VFM}} \in \mathbb{R}^{h \times N\times D_{h}}$.
Specifically, we treat the VFM attention adjacency ${A}_\text{VFM} = {K}_{\text{VFM}} {K}_{\text{VFM}}^\top \in \mathbb{R}^{h \times N \times N}$ as a graph, where $h$ represents the number of heads, and transfer crucial graph patterns from ${A}_\text{VFM}$ to ${A}_\text{CLIP}$.
However, a key question remains: ``How do we account for the \textit{multi-head attention mechanisms} in both VFM and CLIP?" 
In other words, ``how do we match $i$-th VFM attention head ${A}_\text{VFM}^i$ with $j$-th CLIP attention head ${A}_\text{CLIP}^j$?"
A straightforward approach is to distill the attention heads sequentially (i.e., $i=j$). 
Nevertheless, recent studies~\cite{zhu2023weaktr,li2023does,kang2025see} indicate that different attention heads focus on separate parts of the image.
Thus, matching the optimal heads between VFM and CLIP is essential for precise attention distillation.

\subsubsection{Complementary Spectral Graph Matching}
\begin{figure}[t!]
  \centering
   \includegraphics[width=\linewidth]{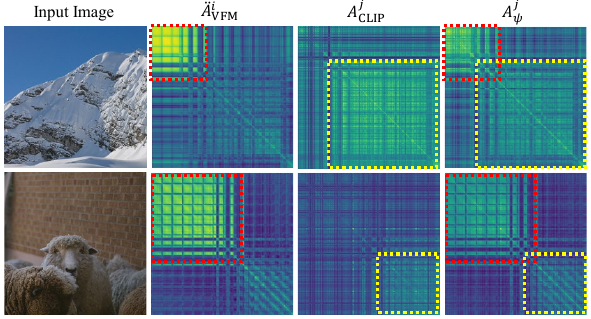}
    \vspace{-20pt}

   \caption{
   Visualization of the matched complementary graph pairs. The red dotted line indicates the focus areas of VFM, while the yellow dotted line highlights those for CLIP.
    By aggregating graphs with contrasting structural properties, we enhance the object-level context in ${A}_\text{CLIP}$ supported by $\ddot{ {A}}_\text{VFM}$, resulting $ {A}_\psi$.
   }
    \vspace{-10pt}
   \label{fig:graph}
\end{figure}

\vspace{-2pt}
\label{sec:method_GM}
We match graphs with \textit{contrasting structures}, as shown in Fig~\ref{fig:graph}, enabling VFM to supplement the object-level contextual knowledge that CLIP alone struggles to capture.
To match an optimal head-wise attention graph, we exploit the spectral field of both ${A}_\text{VFM}$ and ${A}_\text{CLIP}$. 
Our graph-matching mechanism has two main steps as illustrated in Fig.~\ref{fig:local}(a): (I) examining the eigenvalues of each graph and (II) matching optimal graph pairs using the spectral distribution.

\myparagraph{(I) Eigenvalue Examination.} 
For each graph from VFM and CLIP, denoted as ${A}_\text{VFM}^i $ and ${A}_\text{CLIP}^i $ for ($ i = 1, \dots, h $), we perform eigendecomposition. 
This yields the eigenvalues for each head $h$, $\lambda_\text{VFM}^i $ and $\lambda_\text{CLIP}^i$, from which the top $m$ fixed eigenvalues are selected.
These eigenvalues contain each graph's unique structural features, key properties essential for distinguishing different attention graph patterns.

\myparagraph{(II) Graph Matching via Spectral Distribution.} 
After obtaining eigenvalues for each head, we compute spectral distances to quantify structural differences, creating a cost matrix $\mathcal{C} \in \mathbb{R}^{h \times h}$ for each graph pair from VFM and CLIP:
\vspace{-10pt}
\begin{equation}
\mathcal{C}_{ij} = 1 - \mathcal{D}_W(\bar{\lambda}_\text{VFM}^i, \bar{\lambda}_\text{CLIP}^j), \quad \forall \, i, j \in \{1, \dots, h\}
\end{equation}
where $\bar{\lambda}_\text{VFM}^i$ and $\bar{\lambda}_\text{CLIP}^j$ represent the normalized eigenvalues of the $i$-th head of VFM and the $j$-th head of CLIP, respectively, and $\mathcal{D}_W$ denotes the Wasserstein distance.
Here, the Wasserstein distance is computed as $\mathcal{D}_W(u, v) = \sum_{i=1}^h \left| \text{sort}(u)_i - \text{sort}(v)_i\right|$, where $u$ and $v$ are the two input distributions, and the sort function orders the values of $u$ and $v$ in ascending order.
After computing $\mathcal{C}$, we apply the Hungarian matching algorithm~\cite{kuhn1955hungarian} to the cost matrix to find the optimal pairing of graph heads.
This approach pairs graphs with contrasting characteristics, enabling complementary distillation of object-level context knowledge from $ {A}_\text{VFM}$ to $ {A}_\text{CLIP}$, as described in the following section.

\subsubsection{Distilling VFM Spectral Graph to CLIP} 
\vspace{-2pt}
\label{sec:method_lr}
Once we match optimal graph pairs, we distill the VFM graph to CLIP (see Fig.~\ref{fig:local}(b)).
For the sake of explanation, we assume that the $i$-th VFM attention head ${A}_\text{VFM}^i$ is matched with the $j$-th CLIP attention head ${A}_\text{CLIP}^j$.
An intuitive approach would be simply aggregating $ {A}_\text{VFM}^i$ and $ {A}_\text{CLIP}^j$ without any transformation. 
However, as shown in Fig.~\ref{fig:attn_vis} (${A}_{\psi\text{ \textit{w/o} LR}}$), this approach may transfer noise or irrelevant information, highlighting the need to extract features that emphasize object-level context.
From this realization, we leverage the low-rank components of VFM, which contain distinct object patterns within the graph structure~\cite{chung1997spectral}.
Specifically, we (I) extract the critical object-level contextual structure of ${A}_\text{VFM}^i$ via low-rank approximation and enhance the graph structure by dynamically scaling eigenvalues.
Then, we (II) distill the approximated VFM graph into CLIP by aggregating graphs based on spectral distance.

\begingroup
\renewcommand{\arraystretch}{0.95}
\begin{table*}[t]
  \centering
  \caption{Quantitative results with state-of-the-art unsupervised open-vocabulary semantic segmentation models on eight datasets. 
  }
  \vspace{-5pt}
  \setlength{\tabcolsep}{0.4em}
  \resizebox{\textwidth}{!}{
  \begin{tabular}{lccccccccccccc}
    \toprule
    {Model} &  & {Supporting Dataset} & {Extra-Training} & {Fair} & {V21} & {PC60} & {C-Obj} & {V20} & {PC59} & {C-Stf} & {City} & {ADE} & {Avg.} \\
    \midrule
    \rowcolor{gray!5}
    GroupViT~\cite{xu2022groupvit} & \textcolor{gray}{\scriptsize{CVPR'22}} & CC12M+RedCaps & \ding{51} & \ding{55} & 50.4 & 18.7 & 27.5 & 79.7 & 23.4 & 15.3 & 11.1 & 9.2 & 29.4\\
    TCL~\cite{cha2023learning} & \textcolor{gray}{\scriptsize{CVPR'23}} & CC3M+CC12M & \ding{51} & \ding{55} & 55.0 & 30.4 & 31.6 & 83.2 & 33.9 & 22.4 & 24.0 & 17.1 & 37.2\\
    \rowcolor{gray!5}
    CoDe~\cite{wu2024image} & \textcolor{gray}{\scriptsize{CVPR'24}} & CC3M+RedCaps & \ding{51} & \ding{55} & 57.5 & 30.5 & 32.3 & - & - & 23.9 & 28.9 & 17.7 & -\\
    CLIP-DINOiser~\cite{wysoczanska2023clip} & \textcolor{gray}{\scriptsize{ECCV'24}} & ImageNet1k & \ding{51} & \ding{55} & 62.1 & 32.4 & 34.8 & 80.9 & 35.9 & 24.6 & 31.7 & 20.0 & 40.3\\

    \midrule
    \rowcolor{gray!5}
    ReCo~\cite{shin2022reco} & \textcolor{gray}{\scriptsize{NeurIPS'22}} & ImageNet1k & \ding{55} & \ding{55} & 25.1 & 19.9 & 15.7 & 57.7 & 22.3 & 14.8 & 21.6 & 11.2 & 23.5\\
    FOSSIL~\cite{Barsellotti_2024_WACV} & \textcolor{gray}{\scriptsize{WACV'24}} & COCO Captions & \ding{55} & \ding{55} & - & - & - & - & 35.8 & 24.8 & 23.2 & 18.8 & -\\
    \rowcolor{gray!5}
    FreeDa~\cite{barsellotti2024training} & \textcolor{gray}{\scriptsize{CVPR'24}} & COCO Captions & \ding{55} & \ding{55} & - & - & - & 85.6 & 43.1 & 27.8 & 36.7 & 22.4 & -\\
    
    \midrule
    \midrule
    CLIP~\cite{radford2021learning} & \textcolor{gray}{\scriptsize{ICML'21}} & \ding{55} & \ding{55} & \ding{51} & 18.6 & 7.8 & 6.5 & 49.1 & 11.2 & 7.2 & 6.7 & 3.2 & 13.8\\
    \rowcolor{gray!5}
    MaskCLIP~\cite{zhou2022extract} & \textcolor{gray}{\scriptsize{ECCV'22}} & \ding{55} & \ding{55} & \ding{51} & 38.3 & 23.6 & 20.6 & 74.9 & 26.4 & 16.4 & 12.6 & 9.8 & 27.9\\
    GEM~\cite{bousselham2024grounding} & \textcolor{gray}{\scriptsize{CVPR'24}} & \ding{55} & \ding{55} & \ding{51} & 46.2 & - & - & - & 32.6 & - & - & 15.7 & -\\
    \rowcolor{gray!5}
    CaR~\cite{sun2024clip} & \textcolor{gray}{\scriptsize{CVPR'24}} & \ding{55} & \ding{55} & \ding{51} & 48.6 & 13.6 & 15.4 & 73.7 & 18.4 & - & - & 5.4 & -\\
    PnP-OVSS~\cite{luo2024emergent} & \textcolor{gray}{\scriptsize{CVPR'24}} & \ding{55} & \ding{55} & \ding{51} & - & - & 36.2 & 51.3 & 28.0 & 17.9 & - & 14.2 & -\\
    \rowcolor{gray!5}
    CLIPtrase~\cite{shao2024explore} & \textcolor{gray}{\scriptsize{ECCV'24}} & \ding{55} & \ding{55} & \ding{51} & 50.9 & 29.9 & \textcolor{customblue}{43.6} & 81.0 & 33.8 & 22.8 & 21.3 & 16.4 & 32.7\\
    ClearCLIP~\cite{lan2024clearclip} & \textcolor{gray}{\scriptsize{ECCV'24}} & \ding{55} & \ding{55} & \ding{51} & 51.8 & 32.6 & 33.0 & 80.9 & 35.9 & 23.9 & 30.0 & 16.7 & 38.1\\
    \rowcolor{gray!5}
    SCLIP~\cite{wang2023sclip} & \textcolor{gray}{\scriptsize{ECCV'24}} & \ding{55} & \ding{55} & \ding{51} & 59.1 & 30.4 & 30.5 & 80.4 & 34.1 & 22.4 & 32.2 & 16.1 & 38.2\\
    LaVG~\cite{kang2024defense} & \textcolor{gray}{\scriptsize{ECCV'24}} & \ding{55} & \ding{55} & \ding{51} & {62.1} & 31.6 & 34.2 & {82.5} & 34.7 & 23.2 & 26.2 & 15.8 & 38.8\\
    \rowcolor{gray!5}
    ProxyCLIP~\cite{lan2024proxyclip} & \textcolor{gray}{\scriptsize{ECCV'24}} & \ding{55} & \ding{55} & \ding{51} & 59.1 & {35.2} & 36.2 & 78.2 & {38.8} & {26.2} & {38.1} & {19.6} & {41.4}\\
    NACLIP~\cite{hajimiri2025naclip} & \textcolor{gray}{\scriptsize{WACV'25}} & \ding{55} & \ding{55} & \ding{51} & 58.9 & 32.2 & 33.2 & 79.7 & 35.2 & 23.3 & 35.5 & 17.4 & 39.4\\
    \midrule
    \rowcolor{blue!3}
    {\ourmethod} &  & \ding{55} & \ding{55} & \ding{51} & \color{customblue}{65.8} & \textcolor{customblue}{36.7} & {37.8} & \textcolor{customblue}{87.8} & \textcolor{customblue}{40.2} & \textcolor{customblue}{26.7} & \textcolor{customblue}{39.4} &\textcolor{customblue}{20.4} & \textcolor{customblue}{44.4}\\
\bottomrule
  \end{tabular}
  }
  \label{tab:model_comparison}
  \vspace{-10 pt}
\end{table*}
\endgroup

\myparagraph{(I) Low-Rank Dynamic Eigenscaling.} 
Our goal is to transfer the essential object-level contextual structure of VFM graph ${A}_\text{VFM}$ to CLIP. 
To achieve this, we extract low-rank components of the VFM graph using standard eigendecomposition, as $ {A}_\text{VFM}^i = U \Sigma U^\top$, where $U$ and $\Sigma$ represent the eigenvector and the diagonal eigenvalue matrix, respectively.
In the decomposed eigenbasis, we identify key object-level features of each graph by searching an optimal number of eigenvalues $k$ through an energy-based approach. 
This ensures that the chosen $k$ eigenvalues capture a significant portion of the graph's energy, retaining essential structural information while discarding noise and less relevant details.
More details on searching $k$ can be found in the supplementary material.
This process yields a low-rank approximated VFM graph, expressed as $\tilde{ {A}}_\text{VFM}^i = U_k \Sigma_k U_k^\top$.

We refine the low-rank components with a scaling function $\phi$, which dynamically amplifies larger eigenvalues and reduces smaller ones.
Compared to the conventional shrinkage function~\cite{boas2016shrinkage,barash2017optimal}, which only focuses on noise cutoff, our approach emphasizes essential structural information, particularly object-level context features, while suppressing noise and irrelevant details. 
More details on dynamic eigenscaling can be found in the supplementary material. 
The graph after applying dynamic eigenscaling is as
\vspace{-1pt}
\begin{equation}
\ddot{{A}}_\text{VFM}^i = U_k \phi(\Sigma_k) U_k^\top.
\label{eq:des}
\vspace{-1pt}
\end{equation}
\myparagraph{(II) VFM Graph Distillation.} 
We distill the essential object-level contextual knowledge from the tailored VFM graph $\ddot{ {A}}_\text{VFM}^i$ into the $ {A}_\text{CLIP}^j$ by aggregating their structures:
\vspace{-1pt}
\begin{equation}
  {A}_{\psi}^j = (w_{ij} \ddot{ {A}}_\text{VFM}^i +  {A}_{\text{CLIP}}^j)/({w_{ij} + 1}), 
 \vspace{-1pt}
\label{eq:distillation}
\end{equation}
where $w_{ij}$ is defined as the Wasserstein distance between eigenvalues from both graphs as $\mathcal{D}_\mathcal{W}(\bar{\lambda}_\text{VFM}^i, \bar{\lambda}_\text{CLIP}^j)$. This adaptively assigns higher weights to graphs with complementary structures, while reducing influence on those with similar structures.
Our distillation strategy effectively transfers the critical object-level context from VFM to CLIP, ensuring multiple components of the same object are grouped into a unified semantic as illustrated in Fig.~\ref{fig:attn_vis} ($ {A}_{\psi}$).
Therefore, we use $A_\psi$ as our attention matrix instead of $A_\text{CLIP}$ in Eq.~\eqref{eq:softmax} to compute the final visual feature $F_\text{CLIP}$.

\begin{figure}[t!]
  \centering
   \includegraphics[width=\linewidth]{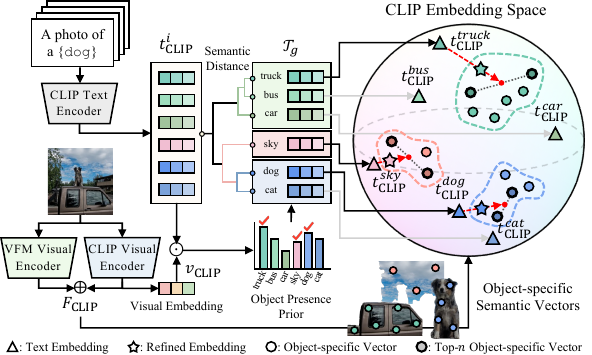}
   \vspace{-20pt}
   \caption{
   Detailed illustration of our object presence prior-guided text embedding adjustment module.
   The CLIP text encoder generates text embeddings for each object class, and the object presence prior is derived from both visual and text embeddings. 
   Within hierarchically defined class groups, text embeddings are selected based on object presence prior, then refined in an object-specific direction to align with components likely present in the image.
   }
   \label{fig:cta}
   \vspace{-10pt}
\end{figure}

\subsection{Object Presence-Driven Object-Level Context}
\label{sec:method_global}
The spectral object-level context distillation in Sec.~\ref{sec:method_local} enables precise object mask generation.
However, due to the nature of OVSS, where users can provide arbitrary query prompts, different parts of the same object may still be assigned to closely related categories.
Meanwhile, numerous studies~\cite{radford2021learning,jiang2023clip,mao2023clip4hoi} have demonstrated that CLIP excels in zero-shot object classification capability.
Accordingly, we utilize the zero-shot object classification score encoded by CLIP (i.e., the object presence prior) to refine text embeddings and patch-text similarity, enhancing object-centered perspective.
To compute the object presence prior $P$, we take an entire input image $\mathcal{I}$ and compute likelihood for each class $i$ as $P(i) = {\{ t_\text{CLIP}^i }\}_{i=1}^C v_\text{CLIP}$, where the visual embedding vector $v_\text{CLIP} = \mathcal{F}_\text{CLIP}^v(\mathcal{I}) \in \mathbb{R}^{d}$ is derived from the ViT \texttt{[CLS]} token.
The object presence prior is used in the following sections to enhance object-level context.

\myparagraph{Object-Guided Text Embedding Adjustment.}
We adjust text embeddings ${\{ t_\text{CLIP}^i }\}_{i=1}^C$ to embed object-specific contextual semantics, as shown in Fig.~\ref{fig:cta}.
To prevent object parts from being assigned to ambiguous classes, we hierarchically cluster text embeddings based on CLIP semantic distance, forming groups of similar embeddings $\mathcal{T}_g = \{ t_\text{CLIP}^i \mid i \in {I}_g \}$, where $\mathcal{T}_g$ is a subset of entire text embeddings ${\{ t_\text{CLIP}^i }\}_{i=1}^C$ with index set ${I}_g$.
Further details about this clustering process are provided in the supplementary materials.
Within $\mathcal{T}_g$, we identify the object most likely to appear in the image using object presence prior $P$, formulated as $i^* = \arg\max_{i \in I_g} P(i)$.
This facilitates refining the text embedding vector $t_\text{CLIP}^{i^*}$ to better represent object-contextual semantics.
We note that patch-wise visual feature ${F}_\text{CLIP} \in \mathbb{R}^{N \times d}$ contains object-specific semantic vectors that can guide $t_\text{CLIP}^{i^*}$ accordingly.
Thus, we compute cosine similarity between ${F}_\text{CLIP}$ and $t_\text{CLIP}^{i^*}$ and select top-$n$ object-specific image vector denoted as $\{{f}_i\}_{i=1}^n := \text{Concat}\left[f_1, \dots, f_n \right] \in \mathbb{R}^{n \times d}$.
To define a targeted direction, we compute the average of $\{{f}_i\}_{i=1}^n$ as $\mu_n \in \mathbb{R}^{d}$, and guide selected text embedding $t_\text{CLIP}^{i^*}$ towards ${\mu}_n$ as
\vspace{-2pt}
\begin{equation}
\Tilde{t}_\text{CLIP}^{i^*} = (1 - \alpha) \cdot {t_\text{CLIP}^{i^*}} + \alpha \cdot {\mu}_n,
\end{equation}
where $\alpha$ controls the degree of alignment with $\mu_n$. The adjusted text embedding $\Tilde{t}_\text{CLIP}^{i^*}$ replaces the original text embedding ${t}_\text{CLIP}^{i}$ in the set ${\{ t_\text{CLIP}^i \}}_{i=1}^C$, which is then used to compute the patch-text similarity logit $\hat{\mathcal{S}}$ in Eq.~\eqref{eq:logit}.

\begin{figure*}[t!]
    \begin{center}
        \includegraphics[width=\textwidth]{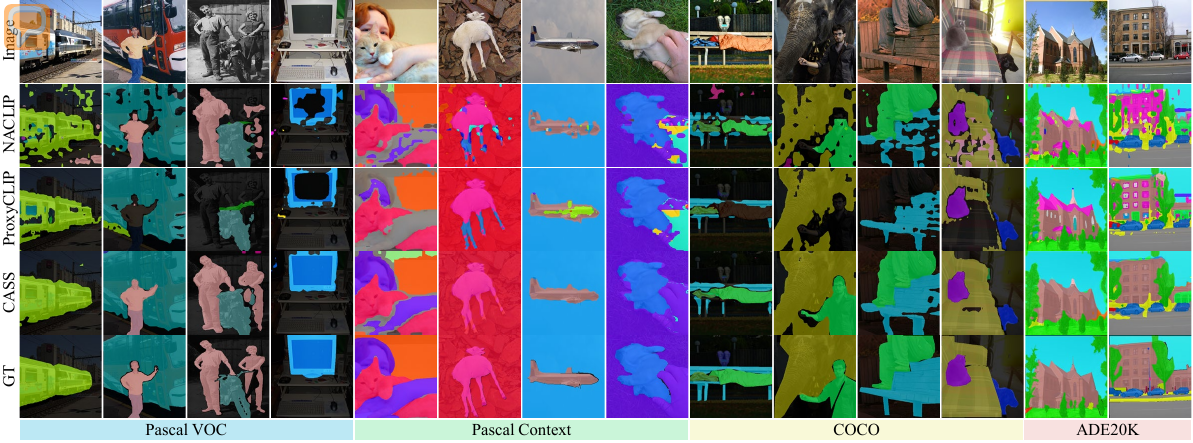}
    \end{center}
    \vspace{-20pt}
    \caption{
    Qualitative comparison across the Pascal VOC~\cite{pascal-voc-2012}, Pascal Context~\cite{mottaghi2014role}, COCO~\cite{caesar2018coco}, and ADE20K~\cite{zhou2019semantic} datasets using CLIP ViT-B/16~\cite{radford2021learning}, evaluating previous state-of-the-art training-free OVSS methods, including NACLIP~\cite{hajimiri2025naclip}, ProxyCLIP~\cite{lan2024proxyclip}, and ours.
    }
    \label{fig:qual}
    \vspace{-12pt}
\end{figure*}

\vspace{1pt}
\myparagraph{Object Perspective Patch-Text Similarity.}
We integrate the object presence prior with patch-text similarity $\hat{\mathcal{S}}$ from Eq.~\eqref{eq:logit} to produce a segmentation map from an object-perspective view. Unlike Barsellotti \textit{et al.}~\cite{barsellotti2024training}, which computes likelihood from an image window $\hat{\mathcal{I}}$, our approach derives similarity from the full image $\mathcal{I}$, enabling richer object-contextual understanding at the broader level.
Thus, we refine the patch-text similarity $\hat{\mathcal{S}}$ from Eq.~\eqref{eq:logit} to obtain the final patch-text similarity $\hat{\mathcal{S}}^*$, computed as
\vspace{-5pt}
\begin{equation}
\hat{\mathcal{S}}^* = (1 - \gamma) \cdot  \hat{\mathcal{S}} + \gamma \cdot   \left\{ t_\text{CLIP}^i \right\}_{i=1}^C{}v_{\text{CLIP}} ,
\vspace{-5pt}
\end{equation}
where $\gamma$ controls the balance between $\hat{\mathcal{S}}$ and object presence prior. 
This allows the final segmentation map to capture object-level details for precise, class-aware segmentation.
\vspace{-5pt}
\section{Experiments}
\label{sec:experiments}

We first describe the experimental settings, including implementation details, datasets, and evaluation metrics (Sec.~\ref{sec:experimental_setting}). We then present quantitative and qualitative evaluation results (Sec.~\ref{sec:eval}) and demonstrate ablation studies to assess each proposed component (Sec.~\ref{sec:ablation}).
Additional experiment results are provided in the supplementary material.

\subsection{Experimental Settings}
\label{sec:experimental_setting}
\vspace{-5pt}
\myparagraph{Implementation Details.}
Similar to existing training-free approaches~\cite{wang2023sclip,hajimiri2025naclip,lan2024proxyclip,kang2024defense,shao2024explore}, we employ CLIP ViT-B/16~\cite{radford2021learning} as our VLM and DINO ViT-B/8~\cite{caron2021emerging} as our VFM, keeping both frozen.
Input images are matched with the shorter side set to 336 pixels (or 560 pixels for the higher-resolution Cityscapes dataset), and sliding window inference is applied with a 224 $\times$ 224 window and a stride of 112 to achieve efficient evaluation coverage.
All experiments are conducted using NVIDIA RTX A6000 Ada$^*$.

\let\thefootnote\relax\footnote{\scriptsize{ 
* Advanced Database System Infrastructure (NFEC-2024-11-300458)
}}

\vspace{-15pt}
\myparagraph{Datasets.}
Our primary datasets include PASCAL VOC 2012~\cite{pascal-voc-2012}, PASCAL Context~\cite{mottaghi2014role}, and COCO~\cite{caesar2018coco}, each evaluated with and without a background class. We refer to the versions with a background class as V21, PC60, and C-Obj, and those without as V20, PC59, and C-Stf. 
We also report results on the ADE20K (ADE)~\cite{zhou2019semantic} and Cityscapes (City)~\cite{cordts2016cityscapes} datasets. 
These datasets contain 21, 60, 81, 20, 59, 171, 150, and 19 classes, respectively.

\subsection{Evaluation Results}
\label{sec:eval}
\vspace{-3pt}
We evaluate our model against training-free OVSS methods both quantitatively and qualitatively. 
For fair comparison, we refrain from using the mask refinement (e.g., PAMR~\cite{pamr} or DenseCRF~\cite{krahenbuhl2011efficient}) and reproduce results from~\cite{lan2024proxyclip} using a $224 \times 224$ window to match the baselines. 
Performance is measured primarily with mean IoU (mIoU), complemented by pixel accuracy (pAcc) for a broader assessment.

\vspace{-3pt}
\myparagraph{Quantitative Evaluation: mIoU.}
Table~\ref{tab:model_comparison} presents the main results in mIoU, where all results are obtained using CLIP ViT-B/16 as the backbone model for fair comparisons.
To demonstrate the strengths of our model, we also include methods that use supporting datasets (i.e., retrieving additional datasets during inference) or additional training (models evaluated under the same setting as ours are indicated with \ding{51} in the ``Fair" column).
Our model, {\ourmethod}, performs favorably against the state-of-the-art methods, with an average gain of \textnormal{3.0} mIoU point across eight datasets.
Notably, for the V20, {\ourmethod} achieves a performance gain of \textnormal{5.3} mIoU point over the second-best model. 
Further, {\ourmethod} outperforms CLIP-DINOiser~\cite{wysoczanska2023clip} (which also leverages DINO with extra data and training) by \textnormal{4.1} mIoU point.

\vspace{-3pt}
\myparagraph{Quantitative Evaluation: pAcc.} 
Table~\ref{tab:aacc} shows the results of pAcc on recent state-of-the-art models evaluated under the same settings. 
Our {\ourmethod} outperforms the second-highest model by \textnormal{2.2} pAcc point  and exceeds one of the most recent papers, CLIPtrase~\cite{shao2024explore}, by \textnormal{9.8} pAcc point.
Our model consistently performs better than existing methods on mIoU and pAcc by effectively capturing object-level context, resulting in more accurate object masking.

\vspace{-3pt}
\myparagraph{Qualitative Analysis.}
Fig.~\ref{fig:qual} provides a qualitative comparison with recent state-of-the-art models~\cite{hajimiri2025naclip,lan2024proxyclip}. 
Existing methods often produce noisy segmentation maps and do not group object components correctly, such as wheels within a car or a person’s head or arm. 
Additionally, they often misclassify object parts into ambiguous classes (e.g., a motorcycle as a bicycle or an airplane as a car). 
In contrast, based on a deep understanding of object-level context, our {\ourmethod} generates clean segmentation maps that accurately group all object components, including slender parts such as human arms or animal legs, into their respective object classes.

\begingroup
\renewcommand{\arraystretch}{1.1}
\begin{table}[t]
  \centering
   \caption{Quantitative results using average pixel accuracy.
   }
   \vspace{-10pt}
  \resizebox{\columnwidth}{!}{
  \begin{tabular}{lccccccccc}
    \toprule
    {Model} &    {V21} & {PC60} & {C-Obj} & {V20} & {PC59} & {C-Stf} & {City} & {ADE} & {Avg.} \\
    \midrule
    CLIPtrase~\cite{shao2024explore}  & 78.6 & 52.1 & 50.1 & 89.7 & 58.9 & 38.9 & 63.4 & 38.6 & 59.1\\
    LaVG~\cite{kang2024defense}  &  {89.3} & 48.7 & 74.8 & {91.1} & 58.9 & 39.1 & 68.5 & 37.0 & 63.4\\
    SCLIP~\cite{wang2023sclip} &  87.6 & 49.2 & 74.3 & 91.0 & 58.3 & 38.4 & 72.7 & 38.7 & 63.8\\
    ProxyCLIP~\cite{lan2024proxyclip}  &  86.6 & {52.0} & {75.9} & 88.4 & {63.4} & {43.4} & {74.9} & \textcolor{customblue}{49.1} & {66.7}\\
    NACLIP~\cite{hajimiri2025naclip}  & 87.1 & 51.2 & 75.3 & 89.2 & 59.8 & 39.3 & 71.7 & 45.2 & 64.9\\
    \midrule
    \rowcolor{blue!3}
    {\ourmethod}   &\textcolor{customblue}{90.1} &\textcolor{customblue}{55.6} & \textcolor{customblue}{76.0} & \textcolor{customblue}{93.9} & \textcolor{customblue}{65.0} & \textcolor{customblue}{43.6} & \textcolor{customblue}{78.2} & 48.6 & \textcolor{customblue}{68.9}\\
\bottomrule
  \end{tabular}
  }
  \label{tab:aacc}
  \vspace{-15 pt}
\end{table}
\endgroup

\vspace{-3pt}
\subsection{Ablation Study}
\label{sec:ablation}
\vspace{-3pt}
We perform ablation studies to analyze the proposed method. 
We use the datasets representative of segmentation tasks for all experiments: V21, PC59, C-Stf, following~\cite{wang2023sclip}.
Please refer to the supplementary material for the additional ablation studies including VFM backbones and feature type.

\myparagraph{Effect of Spectral VFM Distillation.}
Table~\ref{tab:ablation} shows the results of incorporating VFM (Exp.\#3) with the vanilla setting (Exp.\#1), which directly uses the output from Eq.~\eqref{eq:logit}. 
In Exp.\#3, ${A}_\text{VFM}$ and ${A}_\text{CLIP}$ are aggregated in a straightforward head-wise manner, without any additional processing to ${A}_\text{VFM}$.
Simply leveraging ${A}_\text{VFM}$ leads to a performance gain compared to the vanilla setting.
Exp.\#4 to Exp.\#6 validate the effectiveness of the proposed components (see Sec.~\ref{sec:method_local}). 
Exp.\#4 demonstrates the aggregation of ${A}_\text{VFM}$ and ${A}_\text{CLIP}$ using a complementary spectral graph matching (GM) procedure. 
Exp.\#5, denoted as LR, presents the results with only the low-rank approximation applied to ${A}_\text{VFM}$, while Exp.\#6 shows the results with both the low-rank approximation and Dynamic Eigenscaling (DE) applied, as defined in Eq.~\eqref{eq:des}.
Notably, Exp.\#6 shows significant performance gain over simply integrating VFM knowledge (Exp.\#3), demonstrating the effectiveness of graph matching and low-rank approximation in capturing object-level context. 
Fig.~\ref{fig:attn_vis} qualitatively visualizes the attention map, with the left, middle, and right columns corresponding to Exp.\#3, Exp.\#4, and Exp.\#6, respectively.
We observe that integrating proposed low-rank dynamic eigenscaling effectively reduces noise and emphasizes objects by highlighting objects with strong inter-object associations.

\begingroup \small
\setlength{\tabcolsep}{2.9pt} 
\renewcommand{\arraystretch}{0.9} 
\begin{table}[t!]
\caption{Ablation results showing the impact of our components on the performance across V21~\cite{pascal-voc-2012}, PC59~\cite{mottaghi2014role}, and C-Stf~\cite{caesar2018coco}. The abbreviations in the table are noted in Sec.~\ref{sec:ablation}.
}
\vspace{-5pt}
    \centering
    \small 
    \begin{tabular}{cc c ccc c ccccc}
    \hlineB{2.5}
        {\footnotesize Exp.} & \multicolumn{1}{c}{\multirow{2}{*}{${A}_\text{VFM}$}} &   & \multicolumn{1}{c}{\multirow{2}{*}{GM}} & \multicolumn{1}{c}{\multirow{2}{*}{LR}} & \multicolumn{1}{c}{\multirow{2}{*}{DE}} &    & \multicolumn{1}{c}{\multirow{2}{*}{OPS}} & \multicolumn{1}{c}{\multirow{2}{*}{OTA}}  & \multicolumn{3}{c}{\footnotesize Dataset} \\
        
        \# & & & & & & & & & V21 & PC59 & C-Stf \\
    \hlineB{2.5}
        
        1 &     & &  &  &  & &  & &  58.9 & 35.3 & 23.4 \\      
        \rowcolor{gray!5}
        2 &     & &   &  &   & & \checkmark & \checkmark  &  59.2 & 36.5  & 24.0 \\
        3 & \checkmark   & &  &   & & &   &  & 62.9&  37.7 & 25.2 \\
        \rowcolor{gray!5}
        4 & \checkmark   & & \checkmark &  & & &   & &  63.2 & 38.0 & 25.4 \\
        5 & \checkmark   & & \checkmark &  \checkmark &  & &  &  &  64.5 &  38.4 & 25.8 \\
        \rowcolor{gray!5}
        6 & \checkmark   & & \checkmark &\checkmark & \checkmark & & & & 64.8 & 38.6 & 25.9 \\
        7 & \checkmark   & & \checkmark &\checkmark & \checkmark & & \checkmark &   & 65.0  & 39.4 & 26.6 \\
    \hlineB{1.5}
    \rowcolor{blue!3}
        8 & \checkmark   & & \checkmark & \checkmark  & \checkmark & & \checkmark & \checkmark & \textcolor{customblue}{65.8} & \textcolor{customblue}{40.2} & \textcolor{customblue}{26.7}\\
    \hlineB{2.5}
    \end{tabular}
    \vspace{-8pt}
    \label{tab:ablation}
\end{table}
\endgroup

\begin{figure}[t!]
  \centering
   \includegraphics[width=\linewidth]{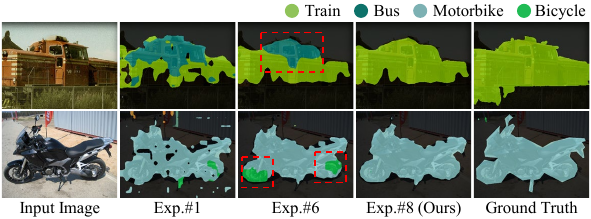}
   \vspace{-18pt}
   \caption{Qualitative ablation of our method. Exp.\# refers to the experiment in Table~\ref{tab:ablation}. 
   As additional components of our proposed method are incorporated, the segmentation map exhibits greater refinement and more effectively captures object-level context.
   }
   \label{fig:ablation}
   \vspace{-14 pt}
\end{figure}

\myparagraph{Effect of Object Presence Prior.}
We validate the effectiveness of the object presence prior-driven methods in Exp.\#7 and Exp.\#8 by examining each component: Object-Perspective Patch-Text Similarity (OPS) and Object-Guided Text Embedding Adjustment (OTA) (see Sec.~\ref{sec:method_global}).
These results show that each component effectively improves performance, emphasizing object-level context.
Additionally, incorporating the object presence prior-driven methods on top of the vanilla setting, as shown in Exp.\#2, clarifies the effectiveness of the proposed method.

\myparagraph{Qualitative Ablations.}
Fig.~\ref{fig:ablation} shows the segmentation maps for the vanilla setting (Exp.\#1), after applying spectral VFM distillation (Exp.\#6), and with all proposed components applied (Exp.\#8). 
While Exp.\#6 improves object masking compared to Exp.\#1, it often misses parts of objects (highlighted by the red dotted line). 
In contrast, incorporating all proposed components (Exp.\#8) consolidates objects into single entities, effectively capturing object-level context in the segmentation mask.

\myparagraph{Different CLIP Backbones.} 
Table~\ref{tab:backbone} shows results on different CLIP backbones, ViT-B/32 and ViT-L/14.
Our {\ourmethod} performs favorably across various backbones, consistently outperforming existing methods in mIoU and pAcc.
These results demonstrate the effectiveness and robustness of our method across different backbone architectures.

\begin{figure}[t]
  \centering
   \includegraphics[width=\linewidth]{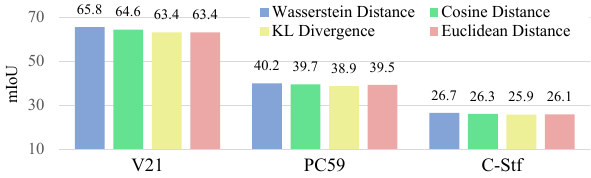}
   \vspace{-20pt}
   \caption{Performance evaluation of different distance metrics used in graph matching and graph distillation.}
   \label{fig:distance}
   \vspace{-8pt}
\end{figure}

\myparagraph{Distance Metric for Graph Matching.}
We evaluate different distance metrics $\mathcal{D}_W$ used in Sec.~\ref{sec:method_GM}, including Wasserstein Distance, Cosine Distance, KL Divergence, and Euclidean Distance.
As shown in Fig.~\ref{fig:distance}, the Wasserstein Distance shows the best results as it provides an intuitive measure of the overall shape of sorted eigenvalue distributions.
While we use Wasserstein Distance due to its performance, the other well-known metrics exhibited comparable results, demonstrating the robustness of our method in terms of the choice of a distance metric and further supporting the effectiveness of our proposed approach.
\let\thefootnote\relax\footnote{\scriptsize{{\bf Acknowledgement.} 
 This work was supported in part by IITP RS-2024-00457882 (National AI Research Hub Project), IITP 2020-0-01361, NRF RS-2024-00345806,  NRF RS-2023-00219019, and RS-2024-00403860 (Korea Basic Science Institute, National Research Facilities and Equipment Center). 
}}

\begingroup
\renewcommand{\arraystretch}{1.0}
\begin{table}[t]
  \centering
   \caption{Ablation results on different CLIP backbones.}
   \vspace{-10pt}
  \resizebox{\columnwidth}{!}{
  \begin{tabular}{c l c c c c c c c c}
    \hlineB{3.5}
    \multirow{2}{*}{\centering\rotatebox{90}{\textbf{ }}} & \multirow{2}{*}{{Model}} & \multicolumn{2}{c}{{V21}} & \multicolumn{2}{c}{{PC59}} & \multicolumn{2}{c}{{C-Stf}} & \multirow{2}{*}{\makecell{{Avg.} \\ {mIoU}}} & \multirow{2}{*}{\makecell{{Avg.} \\ {pAcc}}} \\
    \cmidrule(lr){3-4} \cmidrule(lr){5-6} \cmidrule(lr){7-8}
    & & {mIoU} & {pAcc} & {mIoU} & {pAcc} & {mIoU} & {pAcc} & & \\
    \hlineB{1.5}
    \multirow{5}{*}{\centering\rotatebox{90}{ViT-B/32}} 
    & SCLIP~\cite{wang2023sclip} & 50.6 & 81.0 & 28.7 & 49.6 & 20.0 & 33.4 & 33.1 & 54.7 \\
    & LaVG~\cite{kang2024defense} & 54.8 & 84.4 & 29.0 & 50.1 & 20.5 & 33.7 & 34.8 & 56.1 \\
    & ProxyCLIP~\cite{lan2024proxyclip} & 57.9 & 85.6 & 35.2 & 59.7 & 23.6 & 40.2 & 38.9 & 61.8 \\
    & NACLIP~\cite{hajimiri2025naclip} & 51.1 & 82.8 & 32.4 & 56.8 & 21.2 & 37.1 & 34.9 & 58.9 \\
    & \cellcolor{blue!5} {\ourmethod} & \cellcolor{blue!3}\textcolor{customblue}{58.2} & \cellcolor{blue!3}\textcolor{customblue}{86.2} & \cellcolor{blue!3}\textcolor{customblue}{36.5} & \cellcolor{blue!3}\textcolor{customblue}{61.0} & \cellcolor{blue!3}\textcolor{customblue}{24.4} & \cellcolor{blue!3}\textcolor{customblue}{40.8} & \cellcolor{blue!3}\textcolor{customblue}{39.7} & \cellcolor{blue!3}\textcolor{customblue}{62.7} \\
    \hlineB{1.5}
    \multirow{5}{*}{\centering\rotatebox{90}{ViT-L/14}} 
    & SCLIP~\cite{wang2023sclip} & 44.4 & 78.1 & 25.2 & 46.6 & 17.6 & 29.6 & 29.1 & 51.4 \\
    & LaVG~\cite{kang2024defense} & 51.5 & 84.1 & 27.5 & 50.3 & 19.4 & 32.3 & 32.8 & 55.6 \\
    & ProxyCLIP~\cite{lan2024proxyclip} & 59.8 & 85.7 & 38.3 & 61.0 & 26.2 & \textcolor{customblue}{43.5} & 41.4 & 63.4 \\
    & NACLIP~\cite{hajimiri2025naclip} & 52.2 & 83.1 & 32.1 & 52.8 & 21.4 & 35.7 & 35.2 & 57.2 \\
    & \cellcolor{blue!3} {\ourmethod} & \cellcolor{blue!3}\textcolor{customblue}{62.1} & \cellcolor{blue!5}\textcolor{customblue}{88.0} & \cellcolor{blue!3}\textcolor{customblue}{39.1} & \cellcolor{blue!3}\textcolor{customblue}{62.1} & \cellcolor{blue!3}\textcolor{customblue}{26.3} & \cellcolor{blue!3}43.4 & \cellcolor{blue!3}\textcolor{customblue}{42.5} & \cellcolor{blue!3}\textcolor{customblue}{64.5} \\
    \hlineB{3.5}
  \end{tabular}
  }
  \label{tab:backbone}
  \vspace{-7pt}
\end{table}
\endgroup

\vspace{-5pt}
\section{Conclusion}
\label{sec:conclusion}

In this work, we present {\ourmethod}, a training-free approach for OVSS that integrates object-level context into CLIP. 
By distilling object-aware structural information from VFM into CLIP’s attention, {\ourmethod} effectively enhances intra-object coherence, ensuring that each object is consistently represented as a unified semantic entity. 
Additionally, we leverage object-presence prior knowledge to adjust text embeddings and refine object class assignments, aligning them with object-specific semantic targets in the image.
Consequently, {\ourmethod} effectively groups object components and assigns semantic labels without extra training or additional labeled data, making it a robust solution across diverse datasets and visual contexts.

\clearpage
{
    \small
    \bibliographystyle{ieeenat_fullname}
    \bibliography{main}
}

\clearpage
\setcounter{page}{1}
\maketitlesupplementary

\section{Additional Material: Project Page \& Presentation Video}
We have described our results in an easily accessible manner on our project page, where a brief \textbf{presentation video} is also available. The link to the \textbf{project page} is as follows: \url{https://micv-yonsei.github.io/cass/}.

\section{Detailed Method}
\subsection{Energy-based Low-rank Approximation}
\begin{center}
\begin{minipage}{\linewidth}
\begin{algorithm}[H]
\caption{Optimal Rank $k$ Selection with Low-Rank Eigendecomposition}
\label{alg:optimal_k_lowrank_eigen}
\KwIn{Adjacency matrix ${A} \in \mathbb{R}^{n \times n}$, energy threshold $\eta$, initial rank $q_0$, step size $\Delta q$}
\KwOut{Optimal rank $k$, eigenvectors ${U}$, eigenvalues ${\Sigma}$}

Set $q_{\text{max}} \gets n$ \tcp*[r]{Maximum allowable rank for symmetric matrices}
Set $q \gets q_0$\;

\If{$\| {A} - {A}^\top \|_F > \epsilon$}{
    \textbf{Error:} Input matrix is not symmetric\;
}

\While{$q \leq q_{\text{max}}$}{
    Approximate ${A}$ with rank-$q$ components:
    \[
    {A} \approx {U}_q \Sigma_q {U}_q^\top,
    \]
    where ${U}_q$ contains the top $q$ eigenvectors and $\Sigma_q = \text{diag}(\lambda_1, \lambda_2, \ldots, \lambda_q)$\;
    
    Compute total energy: $E_{\text{total}} \gets \sum_{i=1}^q \lambda_i$
    
    Compute cumulative energy: $E_{\text{cumulative}}(k) \gets \sum_{i=1}^k \lambda_i$
    
    \If{$\frac{E_{\text{cumulative}}(k)}{E_{\text{total}}} \geq \eta$ for some $k \leq q$}{
        \Return $k, {U}_k, \Sigma_k$\;
    }
    
    Increment $q \gets q + \Delta q$\;
}

Set $q \gets q_{\text{max}}$ \tcp*[r]{Fallback to maximum rank if threshold not met}
\Return $q, {U}_q, \Sigma_q$\;
\label{algo:lowrank}
\end{algorithm}
\end{minipage}
\end{center}
We leverage the Energy-based Low-rank Approximation method, as outlined in Algorithm~\ref{algo:lowrank}, which efficiently captures key object-level contextual features within the VFM graph. The method approximates the $i$-th head attention adjacency graph \({A}_\text{VFM}^i\) using the rank-\( q \) approximation as
\begin{equation}
    {A}_\text{VFM}^i \approx {U}_q \Sigma_q {U}_q^\top,
\end{equation}
where \({U}_q\) contains the top \( q \) eigenvectors, and \(\Sigma_q = \text{diag}(\lambda_1, \lambda_2, \dots, \lambda_q)\) consists of the top \( q \) eigenvalues. This selective focus on the most significant components highlights critical object relationships while discarding noise. By iteratively increasing the rank \( q \), the algorithm efficiently captures the graph’s essential structure with minimal complexity by examining cumulative energy.
Our algorithm evaluates whether the retained energy satisfies the threshold \(\eta\) as
\begin{equation}
    \frac{E_{\text{cumulative}}}{E_{\text{total}}} =\frac{\sum_{i=1}^k \lambda_i}{\sum_{i=1}^q \lambda_i} \geq \eta, \quad k \leq q.
\end{equation}
By adaptively identifying the smallest rank \( k \) that meets this criterion, the method ensures a compact and meaningful representation.
Thus, our low-rank approximated VFM graph is expressed as
\begin{equation}
    \Tilde{A}_\text{VFM}^i = {U}_k \Sigma_k {U}_k^\top.
    \label{eq:low}
\end{equation}
Our proposed combination of selective eigendecomposition, energy-based evaluation, and adaptive rank selection enables the algorithm to efficiently preserve object-level contextual features while minimizing computational costs.
In the next section, we refine the computed low-rank components (i.e., the eigenvalues and eigenbasis), through dynamic eigenscaling. This approach further enhances their transformation into more object-centric representations, enabling improved precision and robustness in feature modeling.

\subsection{Dynamic Eigenscaling}
We propose a dynamic eigenscaling function \(\phi\) to refine the eigenvalues \(\Sigma_k\) from Eq.~\eqref{eq:low}. This function amplifies larger eigenvalues while suppressing smaller ones, highlighting dominant components in the graph structure and reducing the influence of noise. The scaled eigenvalues are computed as
\begin{equation}
\Sigma_k' = \phi(\Sigma_k) = \frac{\Sigma_k - \Sigma_{\text{min}}}{\Sigma_{\text{max}} - \Sigma_{\text{min}}} \cdot \text{scaled\_range} + (2-\epsilon) \cdot \Sigma_{\text{min}},
\end{equation}
where \(\Sigma_{\text{min}} = \min(\Sigma_k)\), \(\Sigma_{\text{max}} = \max(\Sigma_k)\), and \(\text{scaled\_range} = \epsilon \cdot \Sigma_{\text{max}} - (2-\epsilon) \cdot \Sigma_{\text{min}}\). 
We fix $\epsilon$ into 1.5.
After applying our proposed dynamic eigenscaling function, our final tailored VFM graph is as
\begin{equation}
\ddot{A}_\text{VFM}^i =  {U}_k \Sigma_k' {U}_k^\top  := {U}_k \phi(\Sigma_k) {U}_k^\top.
\end{equation}
This transformation dynamically adjusts the eigenvalues to retain the graph's most significant features, enhancing object-level contextual representation while minimizing the impact of less relevant components.

\subsection{Hierarchical Grouping of Text Embeddings}

\begin{center}
\begin{minipage}{\linewidth}
\begin{algorithm}[H]
\caption{\small{Pseudo-code of hierarchical clustering in Python style.}}
\definecolor{codeblue}{rgb}{0.25,0.5,0.5}
\definecolor{codegreen}{rgb}{0,0.6,0}
\definecolor{codegray}{rgb}{0.5,0.5,0.5}
\definecolor{codepurple}{rgb}{0.58,0,0.82}
\definecolor{backcolour}{rgb}{0.95,0.95,0.92}
\lstset{
  backgroundcolor=\color{white},
  basicstyle=\fontsize{7.2pt}{7.2pt}\ttfamily\selectfont,
  columns=fullflexible,
  breaklines=true,
  captionpos=b,
  commentstyle=\fontsize{7.2pt}{7.2pt}\color{codeblue},
  numberstyle=\tiny\color{codegray},
  stringstyle=\color{codepurple},
  keywordstyle=\fontsize{7.2pt}{7.2pt}\color{magenta},
}
\begin{lstlisting}[language=python]
from sklearn.metrics.pairwise import cosine_distances
from scipy.cluster.hierarchy import linkage, fcluster
from scipy.spatial.distance import squareform

def hierarchical_clustering(text_embeddings, h_threshold):

    # Compute cosine distance matrix
    distance_matrix = cosine_distances(text_embeddings)
    
    # Convert distance matrix to condensed form
    condensed_matrix = squareform(distance_matrix)
    
    # Perform hierarchical clustering using Ward's method
    Z = linkage(condensed_matrix, method='ward')
    
    # Form flat clusters based on the threshold
    clusters = fcluster(Z, t=h_threshold, criterion='distance') - 1  
    
    return clusters
\end{lstlisting}
\label{algo:hierarchical_clustering_algo}
\end{algorithm}
\end{minipage}
\end{center}
We provide details on grouping class prompts to identify semantically related object class prompts (e.g., \texttt{motorcycle} and \texttt{bicycle}) within the Object-Guided Text Embedding Adjustment module. 
As described in Algorithm~\ref{algo:hierarchical_clustering_algo}, we compute the semantic distance between user-defined text prompts by measuring the cosine distance of their CLIP text embeddings in the CLIP latent space, forming a distance matrix.
Next, the distance matrix is converted into a condensed form to facilitate hierarchical clustering using Ward's method, which minimizes variance within clusters.
Finally, clusters are formed by applying a predefined height threshold to the hierarchical tree, ensuring that semantically similar class prompts are grouped together. 
This enables the identification of class prompt-level relationships that are leveraged to refine the text embeddings towards object-specific semantics.

\section{Additional Experiments}
\subsection{Additional Evaluation Results}
In this section, we provide additional evaluation results including qualitative evaluation (\ref{sec:supp_qul}), real-world open-vocabulary semantic segmentation result (\ref{sec:supp_realworld}), and scale-up version of {\ourmethod} (\ref{sec:supp_scaleup}).

\subsubsection{Additional Qualitative Evaluation}
\label{sec:supp_qul}
We present additional qualitative evaluation results: Fig.~\ref{fig:supp_voc} illustrates results on PASCAL VOC (V20 \& V21)~\cite{pascal-voc-2012}, Fig.~\ref{fig:supp_coco} on COCO (C-Stf \& C-Obj)~\cite{caesar2018coco}, Fig.~\ref{fig:supp_pascal} on PASCAL Context (PC59 \& PC60)~\cite{mottaghi2014role}, and Fig.~\ref{fig:supp_adecity} on both ADE20K (ADE)~\cite{zhou2019semantic} and Cityscapes (City)~\cite{cordts2016cityscapes}.
Note that all the results in this subsection are presented without mask refinement steps, such as PAMR~\cite{araslanov2020single} or DenseCRF~\cite{krahenbuhl2011efficient}, to ensure a fair comparison and to evaluate the capability of the model itself.

\subsubsection{Scale-up Version}
\label{sec:supp_scaleup}

\begingroup
\setlength{\tabcolsep}{6pt} 
\renewcommand{\arraystretch}{1.2} 

\begin{table}[ht]
\centering
\caption{Performance of scale-up version of {\ourmethod}.}
\resizebox{\columnwidth}{!}{
\begin{tabular}{l c c c  c c c c c c c}
\toprule[1.2pt]
{Model} & & \makecell{{Feature} \\ {Encoder}} & \makecell{\texttt{[CLS]} \\ {Encoder}} & {V21} & {PC60} & {C-Obj} & {V20} & {PC59} & {ADE} & {Avg} \\ 
\midrule
CaR~\cite{sun2024clip} & \textcolor{gray}{\scriptsize{CVPR'24}} & ViT-B/16 & ViT-B/16 & 48.6 & 13.6 & 15.4 & 73.7 & 18.4 & 5.4  & 29.2 \\ 
\rowcolor{blue!3}
\ourmethod &      & ViT-B/16 & ViT-B/16 & \textcolor{customblue}{65.8} & \textcolor{customblue}{36.7} & \textcolor{customblue}{37.8} & \textcolor{customblue}{87.8} & \textcolor{customblue}{40.2} & \textcolor{customblue}{20.4} & \textcolor{customblue}{48.1} \\ 
\midrule
\midrule
CaR~\cite{sun2024clip} & \textcolor{gray}{\scriptsize{CVPR'24}}  & ViT-B/16 & ViT-L/14 & 63.7 & 29.4 & 35.1 & \textcolor{customblue}{91.4} & 38.4 & 16.9 & 45.8 \\ 
\rowcolor{blue!3}
\ourmethod     &   & ViT-B/16 & ViT-L/14 & \textcolor{customblue}{66.3} & \textcolor{customblue}{37.0} & \textcolor{customblue}{38.3} & 89.3 & \textcolor{customblue}{40.7} & \textcolor{customblue}{20.7} & \textcolor{customblue}{48.7} \\ 
\bottomrule[1.2pt]
\end{tabular}
}

\label{table:supp_scale}
\end{table}
\endgroup

The zero-shot object classification performance of CLIP~\cite{radford2021learning} improves as the capacity of CLIP increases. 
Unlike the main experiment, where all methods were standardized using CLIP ViT-B/16 for a fair comparison, here we compute the object presence prior using CLIP ViT-L/14 (i.e., the scaled-up version), as reported in Table~\ref{table:supp_scale}.
Since CaR~\cite{sun2024clip} also classifies the proposed mask using CLIP's global image embedding for classification, we compared our method under the same conditions as CaR.
The encoder used for patch-wise dense representation (i.e., $F_\text{CLIP}$ in our main manuscript) is referred to as the Feature Encoder, while the encoder for encoding CLIP visual embedding vector (i.e., $v_\text{CLIP}$) is referred to as the \texttt{[CLS]} Encoder.
The CLIP visual embedding vector serves as an object presence prior in our approach and as a mask proposal classifier in CaR.
For this experiment, we use datasets that are reported in CaR and evaluate with the mIoU metric.
Note that we exclude a mask refinement step (e.g., PAMR~\cite{araslanov2020single} or DenseCRF~\cite{krahenbuhl2011efficient}) for a fair comparison.

When using CLIP ViT-L/14 as the \texttt{[CLS]} encoder in our {\ourmethod}, we observe a noticeable performance improvement compared to the ViT-B/16 configuration. Furthermore, our approach consistently outperforms CaR across all benchmarks under both encoder settings. Notably, when comparing models fairly by using the same \texttt{[CLS]} encoder configuration as CaR (ViT-B/16), CaR experiences a significant drop in performance, particularly on datasets such as PC60 and ADE. In contrast, our method demonstrates robust performance in both settings, maintaining high accuracy and adaptability regardless of the encoder used. 
This highlights the stability and effectiveness of our method, especially in handling variations in encoder configurations, demonstrating its robustness compared to CaR.

\subsection{Additional Ablation Study}
For ablation studies, we use representative datasets for segmentation tasks: PASCAL VOC (V21), PASCAL Context (PC59), and COCO-Stuff (C-Stf), following~\cite{wang2023sclip}.
\subsubsection{Ablation: VFM Backbones}
\begingroup
\setlength{\tabcolsep}{6pt} 
\renewcommand{\arraystretch}{1.2} 
\begin{table}[ht]
  \centering
   \caption{Ablation results with different VFM backbones. For the CLIP visual encoder, ViT-B/16 is used.}
  {
  \resizebox{\columnwidth}{!}{
  \begin{tabular}{c l c c c c c c c c c}
    \hlineB{3.5}
    \multirow{2}{*}{\centering\rotatebox{90}{{ }}} & \multirow{2}{*}{{Model}} & & \multicolumn{2}{c}{{V21}} & \multicolumn{2}{c}{{PC59}} & \multicolumn{2}{c}{{C-Stf}} & \multirow{2}{*}{\makecell{{Avg.} \\ {mIoU}}} & \multirow{2}{*}{\makecell{{Avg.} \\ {pAcc}}} \\
    \cmidrule(lr){4-5} \cmidrule(lr){6-7} \cmidrule(lr){8-9}
    & & & {mIoU} & {pAcc} & {mIoU} & {pAcc} & {mIoU} & {pAcc} & & \\
    \hlineB{1.5}
    \multirow{3}{*}{\makecell{DINOv1 \\ ViT-B/8}}
    & LaVG~\cite{kang2024defense} & \textcolor{gray}{\scriptsize{ECCV'24}} & 62.1 & 89.3 & 34.7 & 58.9 & 23.2 & 39.1 & 40.0 & 62.4 \\
    & ProxyCLIP~\cite{lan2024proxyclip} & \textcolor{gray}{\scriptsize{ECCV'24}} & 59.1 & 86.6 & 38.8 & 63.4 & 26.2 & 43.4 & 41.4 & 64.5 \\
    & \cellcolor{blue!3}{\ourmethod} & \cellcolor{blue!3} & \cellcolor{blue!3}\textcolor{customblue}{65.8} & \cellcolor{blue!3}\textcolor{customblue}{90.1} & \cellcolor{blue!3}\textcolor{customblue}{40.2} & \cellcolor{blue!3}\textcolor{customblue}{65.0} & \cellcolor{blue!3}\textcolor{customblue}{26.7} & \cellcolor{blue!3}\textcolor{customblue}{43.6} & \cellcolor{blue!3}\textcolor{customblue}{44.2} & \cellcolor{blue!3}\textcolor{customblue}{66.2} \\
    \hlineB{1.5}
    \multirow{3}{*}{\makecell{DINOv1 \\ ViT-B/16}}
    & LaVG~\cite{kang2024defense} & \textcolor{gray}{\scriptsize{ECCV'24}} & 61.5 & 88.9 & 34.6 & 58.9 & 22.8 & 38.8 & 39.6 & 62.2 \\
    & ProxyCLIP~\cite{lan2024proxyclip} & \textcolor{gray}{\scriptsize{ECCV'24}} & 56.6 & 85.8 & 37.4 & 62.0 & 25.1 & 42.2 & 39.7 & 63.3 \\
    & \cellcolor{blue!3}{\ourmethod} & \cellcolor{blue!3} & \cellcolor{blue!3}\textcolor{customblue}{64.3} & \cellcolor{blue!3}\textcolor{customblue}{89.1} & \cellcolor{blue!3}\textcolor{customblue} {38.9} & \cellcolor{blue!3}\textcolor{customblue}{63.8} & \cellcolor{blue!3}\textcolor{customblue}{26.1} & \cellcolor{blue!3}\textcolor{customblue}{43.1} & \cellcolor{blue!3}\textcolor{customblue}{43.1} & \cellcolor{blue!3}\textcolor{customblue}{65.3} \\
    \hlineB{1.5}
    \multirow{3}{*}{\makecell{DINOv2 \\ ViT-B/14}}
    & LaVG~\cite{kang2024defense} & \textcolor{gray}{\scriptsize{ECCV'24}} & 25.3 & 75.3 & 25.1 & 49.3 & 17.3 & 34.4 & 22.6 & 53.0 \\
    & ProxyCLIP~\cite{lan2024proxyclip} & \textcolor{gray}{\scriptsize{ECCV'24}} & 57.1 & 85.2 & 37.3 & 61.4 & \textcolor{customblue}{25.3} & \textcolor{customblue}{42.3} & 39.9 & 63.0 \\
    & \cellcolor{blue!3}{\ourmethod} & \cellcolor{blue!3} & \cellcolor{blue!3}\textcolor{customblue}{63.0} & \cellcolor{blue!3}\textcolor{customblue}{88.6} & \cellcolor{blue!3}\textcolor{customblue}{38.1} & \cellcolor{blue!3}\textcolor{customblue}{62.2} & \cellcolor{blue!3}\textcolor{customblue}{25.3} & \cellcolor{blue!3}{42.0} & \cellcolor{blue!3}\textcolor{customblue}{42.1} & \cellcolor{blue!3}\textcolor{customblue}{64.3} \\
    \hlineB{3.5}
  \end{tabular}
  }
  }
  \label{tab:dino_backbone}
\end{table}
\endgroup
Table~\ref{tab:dino_backbone} presents the results with various VFM backbones~\cite{caron2021emerging,oquab2023dinov2}, including DINOv1 ViT-B/8, DINOv1 ViT-B/16, and DINOv2 ViT-B/14. Among these, DINOv1 ViT-B/8 serves as the backbone for our main results. In this experiment, we use ViT-B/16 as the CLIP visual encoder.  
To ensure a consistent and fair comparison, we include LaVG~\cite{kang2024defense} and ProxyCLIP~\cite{lan2024proxyclip} as baselines, as both utilize DINO as their backbone model. Since LaVG does not report performance on DINOv1 ViT-B/16 and DINOv2 ViT-B/14, we report reproduced results. Notably, LaVG employs spectral techniques, particularly graph partitioning, making it directly comparable to our approach, which also builds on spectral methods.  
The results show that our method consistently outperforms the baselines across all backbones, demonstrating its robustness and adaptability to different VFMs. In particular, compared to LaVG, which uses graph partitioning for spectral techniques, our method more effectively leverages spectral methods by distilling low-rank components into CLIP via spectral-based graph matching, offering a more advanced solution for training-free OVSS.

\subsubsection{Ablation: Feature Type}

\begingroup
\setlength{\tabcolsep}{10pt} 
\renewcommand{\arraystretch}{1.0} 

\begin{table}[ht]
\centering
  \centering
   \caption{Comparison of features used for constructing attention graphs.}
  \resizebox{\columnwidth}{!}{
  {
    \begin{tabular}{c c c c c c c c c c}
      \hlineB{3.5}
      \multirow{2}{*}{{VFM}} & \multirow{2}{*}{{CLIP}} & \multicolumn{2}{c}{{V21}} & \multicolumn{2}{c}{{PC59}} & \multicolumn{2}{c}{{C-Stf}} & \multirow{2}{*}{\makecell{{Avg.} \\ {mIoU}}} & \multirow{2}{*}{\makecell{{Avg.} \\ {pAcc}}} \\
      \cmidrule(lr){3-4} \cmidrule(lr){5-6} \cmidrule(lr){7-8}
      & & {mIoU} & {pAcc} & {mIoU} & {pAcc} & {mIoU} & {pAcc} & & \\
      \hlineB{1.5}
      $FF$ & $QQ$ & 51.1 & 81.8 & 33.0 & 57.2 & 21.8 & 39.1 & 35.3 & 59.4 \\
      $FF$ & $QK$ & 44.6 & 77.3 & 30.2 & 54.0 & 19.9 & 37.1 & 31.6 & 56.1 \\
      $FF$ & $KK$ & 60.8 & 87.6 & 37.9 & 62.1 & 25.2 & 42.0 & 41.3 & 63.9 \\
      \midrule
      $QQ$ & $QQ$ & 57.1 & 84.9 & 36.5 & 61.2 & 23.9 & 41.5 & 39.2 & 62.5 \\
      $QQ$ & $QK$ & 50.9 & 53.1 & 33.7 & 43.6 & 21.9 & 30.2 & 35.5 & 42.3 \\
      $QQ$ & $KK$ & 64.9 & 89.3 & 39.7 & 64.5 & 26.4 & 43.5 & 43.7 & 65.8 \\
      \midrule
      $QK$ & $QQ$ & 33.7 & 58.1 & 25.2 & 45.7 & 17.3 & 32.4 & 25.4 & 45.4 \\
      $QK$ & $QK$ & 30.7 & 53.1 & 23.2 & 43.6 & 15.5 & 30.2 & 23.1 & 42.3 \\
      $QK$ & $KK$ & 29.0 & 49.1 & 22.1 & 41.7 & 15.0 & 28.6 & 22.0 & 39.8 \\
      \midrule
      $KK$ & $QQ$ & 64.9 & 89.3 & 40.0 & 64.6 & 26.6 & 43.7 & 44.0 & 65.8 \\
      $KK$ & $QK$ & 60.3 & 87.6 & 38.5 & 63.4 & 25.4 & 43.0 & 41.4 & 64.7 \\
      \rowcolor{blue!3}
      $KK$ & $KK$ & \textcolor{customblue}{65.8} & \textcolor{customblue}{90.1} & \textcolor{customblue}{40.2} & \textcolor{customblue}{65.0} & \textcolor{customblue}{26.7} & \textcolor{customblue}{43.6} & \textcolor{customblue}{44.2} & \textcolor{customblue}{66.2} \\

      \hlineB{3.5}
    \end{tabular}
    }
  }
  \label{tab:supple_feature}
  
\end{table}
\endgroup

Table~\ref{tab:supple_feature} presents the performance impact of various features employed in constructing $A_\psi$ during VFM graph distillation. 
The VFM column represents the features utilized in the construction of $A_\text{VFM}$, while the CLIP column represents the features used for $A_\text{CLIP}$. 
In the table, $F$ refers to visual features derived from VFM, and $Q$ and $K$ represent the attention query and key components from their respective models. 
Each row represents a specific combination of features from VFM and CLIP used to construct the attention graph (e.g., $QK$ in the VFM column indicates that $A_\text{VFM}$ is constructed as $A_\text{VFM} = QK$).

When using visual features $F$ from VFM, the results show moderate performance overall; however, constructing $A_\text{VFM}$ using attention components such as $Q$ or $K$ instead of $F$ resulted in better performance.
This indicates that attention components more effectively capture semantic relationships between patches, demonstrating the efficiency of using attention to extract object-level context, which aligns with the findings of prior study~\cite{eagle2024}.
In terms of attention components, $QQ$ performs better than $QK$ across most datasets but is outperformed by $KK$. 
The $QK$ configuration generally shows weaker performance compared to both $QQ$ and $KK$, highlighting the importance of modifying the last layer of self-attention in CLIP instead of using vanilla attention settings.
Overall, the best performance is achieved when $KK$ is used for the attention graph consistently in both VFM and CLIP, as demonstrated by the highest average mIoU of 44.2 and average pAcc of 66.2. This suggests that the $KK$ provides the most robust and object-centric representation for constructing attention graphs.

\subsubsection{Ablation: Graph Matching Strategy}

\begingroup
\setlength{\tabcolsep}{8pt} 
\renewcommand{\arraystretch}{1.2} 
\begin{table}[ht]
  \centering
   \caption{Ablation results on different graph matching strategies.}
  \resizebox{\columnwidth}{!}{
  {
  \begin{tabular}{l c c c c c c c c}
    \hlineB{3.5}
     \multirow{2}{*}{{Method}} & \multicolumn{2}{c}{{V21}} & \multicolumn{2}{c}{{PC59}} & \multicolumn{2}{c}{{C-Stf}} & \multirow{2}{*}{\makecell{{Avg.} \\ {mIoU}}} & \multirow{2}{*}{\makecell{{Avg.} \\ {pAcc}}} \\
    \cmidrule(lr){2-3} \cmidrule(lr){4-5} \cmidrule(lr){6-7}
    & {mIoU} & {pAcc} & {mIoU} & {pAcc} & {mIoU} & {pAcc} & & \\
    \hlineB{1.5}
    Sequential & 64.9 & 89.6 & 39.6 & 64.5 & {26.2} & {43.1} & 43.5 & 65.7 \\
    \midrule
    Similar & 65.5 & 90.0 & 40.2 & 65.1 & 26.6 & 43.6 & 44.1 & 66.2 \\
    \cellcolor{blue!3}{Complementary} & \cellcolor{blue!3}\textcolor{customblue}{65.8} & \cellcolor{blue!3}\textcolor{customblue}{90.1} & \cellcolor{blue!3}\textcolor{customblue}{40.2} & \cellcolor{blue!3}\textcolor{customblue}{65.0} & \cellcolor{blue!3}\textcolor{customblue}{26.7} & \cellcolor{blue!3}\textcolor{customblue}{43.6} & \cellcolor{blue!3}\textcolor{customblue}{44.2} & \cellcolor{blue!3}\textcolor{customblue}{66.2} \\
    \hlineB{3.5}
  \end{tabular}
  }
  }
  \label{tab:complementary}
  
\end{table}
\endgroup

In Table~\ref{tab:complementary}, we compare different graph matching strategies.
``Sequence" refers to matching graphs head-to-head in a sequential manner without applying additional structural alignment techniques. ``Similar" involves matching graphs that exhibit similar structures between VFM and CLIP. Lastly, ``Complementary" uses a strategy that matches graphs with contrasting but complementary structures between VFM and CLIP to leverage diverse information.
From the results, we observe that the ``Sequential" method delivers competitive performance, achieving an average mIoU of 43.5 and pAcc of 65.7. 
The rows for “Similar” and “Complementary” consider the attention graph structures from each head, matching either similar or dissimilar ones, respectively.
The ``Similar" strategy improves the results, with an average mIoU of 44.1 and pAcc of 66.2. However, the proposed ``Complementary" method outperforms both matching methods, achieving the highest average mIoU of 44.2 and pAcc of 66.2. 
Overall, while sequential approaches are straightforward and offer reasonable performance, these results highlight the advantages of considering in matching attention heads.

\subsubsection{Ablation: Number of Eigenvalues for Low-Rank Components in VFM Graph}

\begingroup
\setlength{\tabcolsep}{6pt} 
\renewcommand{\arraystretch}{1.2} 
\begin{table}[ht]
  \centering
   \caption{Ablation results on rank selection methods for low-rank components.}
  \resizebox{\columnwidth}{!}{
  {
  \begin{tabular}{l c c c c c c c c c}
    \hlineB{3.5}
     \multirow{2}{*}{{Method}} &\multirow{2}{*}{{Adaptive}} & \multicolumn{2}{c}{{V21}} & \multicolumn{2}{c}{{PC59}} & \multicolumn{2}{c}{{C-Stf}} & \multirow{2}{*}{\makecell{{Avg.} \\ {mIoU}}} & \multirow{2}{*}{\makecell{{Avg.} \\ {pAcc}}} \\
    \cmidrule(lr){3-4} \cmidrule(lr){5-6} \cmidrule(lr){7-8}
    & & {mIoU} & {pAcc} & {mIoU} & {pAcc} & {mIoU} & {pAcc} & & \\
    \hlineB{1.5}
    $k=3$ & \ding{55} & 62.7 & 89.3 & 39.2 & 64.0 & 25.9 & 42.6 & 42.6 & 65.3 \\
    $k=4$ & \ding{55} & 64.5 & 89.8 & 39.8 & 64.6 & 26.4 & 43.1 & 43.6 & 65.8 \\
    \midrule
    Eigengap & \ding{51} & 64.9 & 89.6 & 39.8 & 64.6 & 26.3 & 43.3 & 43.7 & 65.8 \\
    \cellcolor{blue!3}{Energy-based} & \cellcolor{blue!3}{\ding{51}} & \cellcolor{blue!3}\textcolor{customblue}{65.8} & \cellcolor{blue!3}\textcolor{customblue}{90.1} & \cellcolor{blue!3}\textcolor{customblue}{40.2} & \cellcolor{blue!3}\textcolor{customblue}{65.0} & \cellcolor{blue!3}\textcolor{customblue}{26.7} & \cellcolor{blue!3}\textcolor{customblue}{43.6} & \cellcolor{blue!3}\textcolor{customblue}{44.2} & \cellcolor{blue!3}\textcolor{customblue}{66.2} \\
    \hlineB{3.5}
  \end{tabular}
  }
  }
  \label{tab:eigenvalues}
  
\end{table}
\endgroup

In this experiment, we compare methods for selecting the rank (number of eigenvalues) for low-rank components.
Table~\ref{tab:eigenvalues} presents the results of applying several well-known methods alongside our proposed energy-based approach. 
Rows marked with \ding{55} represent the use of a fixed rank $k$ across the entire images in datasets, whereas rows marked with \ding{51} indicate searching for the optimal rank on a per-image basis.
The eigengap is computed as the index \(i^*\) where the difference between consecutive eigenvalues in the spectrum is maximized, typically defined as $i^* = \underset{i \in \{1, 2, \dots, n-1\}}{\arg\max} \, (\lambda_i - \lambda_{i+1})$, where \(\lambda_i\) and \(\lambda_{i+1}\) represent the \(i\)-th and \((i+1)\)-th largest eigenvalues, respectively.
This metric is commonly used to identify the point where the eigenvalue spectrum exhibits a significant drop, which helps in determining the optimal rank for low-rank approximations or clustering applications.

Fixed-rank methods maintain consistent ranks across all images but fail to adapt to varying data complexity within individual images.
As a result, their performance is lower compared to adaptive approaches.
The eigengap method, which adaptively selects the rank by identifying a significant drop in the eigenvalue spectrum, improves performance compared to fixed-rank strategies.
However, the eigengap approach is limited by its sensitivity to small variations in the eigenvalue distribution, which can lead to inconsistent rank selections in certain cases.
Our energy-based method further enhances performance by selecting ranks based on the cumulative energy of the eigenvalues, ensuring a more stable and data-driven approach to rank determination. 
This method achieves the best overall results and consistently outperforms all other methods across individual datasets.
These results highlight the importance of adaptive rank selection strategies in capturing the intrinsic object-level structure of graphs. 

\section{Implementation Details}
\subsection{Prompt Templates}
\begin{table}[ht]
\centering
\caption{Examples of the prompt templates used for text embedding generation.}
\begin{tabular}{c}
\toprule
{Examples of Prompt Templates} \\
\midrule
``a photo of my $\{ \texttt{prompt} \}$." \\
``a photo of a nice $\{ \texttt{prompt} \}$." \\
``a cropped photo of a $\{ \texttt{prompt} \}$." \\
``a photo of a large $\{ \texttt{prompt} \}$." \\
``art of the $\{ \texttt{prompt} \}$." \\
\bottomrule
\end{tabular}
\label{tab:prom}
\end{table}

Following recent works~\cite{hajimiri2025naclip, wang2023sclip, kang2024defense}, we employ a comprehensive set of prompt templates to enhance the diversity of text embeddings (e.g., ``a photo of $\{ \texttt{prompt} \}$"). 
Specifically, given a prompt query $t$, it is systematically inserted into a predefined list of 80 diverse templates designed to capture varying linguistic contexts. 
Each generated sentence is processed through the CLIP text encoder, resulting in 80 distinct text embeddings. 
To make these embeddings into a unified representation, we compute their average, yielding a single, contextually rich text embedding $t_\text{CLIP}$ that corresponds to the input query $t$. 
The examples of prompt templates are listed in Table~\ref{tab:prom}.

\subsection{Hyperparameters}

\begingroup
\setlength{\tabcolsep}{6pt} 
\renewcommand{\arraystretch}{1.2} 
\begin{table}[ht]
  \centering
   \caption{Hyperparameters used in {\ourmethod}.}
  \resizebox{\columnwidth}{!}{
  {
  \begin{tabular}{c c c c c c c c c c c c c c c c }
    \toprule
      \multicolumn{2}{c}{{V21}} & \multicolumn{2}{c}{{PC60}} & \multicolumn{2}{c}{{C-Obj}} & \multicolumn{2}{c}{{V20}} & \multicolumn{2}{c}{{PC59}} & \multicolumn{2}{c}{{C-Stf}} & \multicolumn{2}{c}{{City}} & \multicolumn{2}{c}{{ADE}}\\
    \cmidrule(lr){1-2} \cmidrule(lr){3-4} \cmidrule(lr){5-6} \cmidrule(lr){7-8} \cmidrule(lr){9-10} \cmidrule(lr){11-12} \cmidrule(lr){13-14} \cmidrule(lr){15-16}
    $\alpha$ & $\gamma$ & $\alpha$ & $\gamma$ & $\alpha$ & $\gamma$ & $\alpha$ & $\gamma$ & $\alpha$ & $\gamma$ & $\alpha$ & $\gamma$ & $\alpha$ & $\gamma$ & $\alpha$ & $\gamma$\\
    \midrule
     0.03 & 0.10 & 0.04 & 0.25 & 0.01 & 0.25 & 0.02 & 0.40 & 0.04 & 0.25 & 0.02 & 0.20 & 0.03 & 0.10 & 0.05 & 0.30 \\
    \bottomrule
  \end{tabular}
  }
  }
  \label{tab:hyperparameter}
\end{table}
\endgroup
Here, we provide a detailed description of the hyperparameters listed in Table~\ref{tab:hyperparameter}. The parameter $\alpha$ controls the balance between the original text embedding and the object-specific vector, while $\gamma$ adjusts the trade-off between the patch-text similarity $\hat{\mathcal{S}}$ and the object presence prior.

\section{Application}
\subsection{Image Inpainting and Object Removal}

{
\begin{figure*}[ht]
  \centering
   \includegraphics[width=0.8\linewidth]{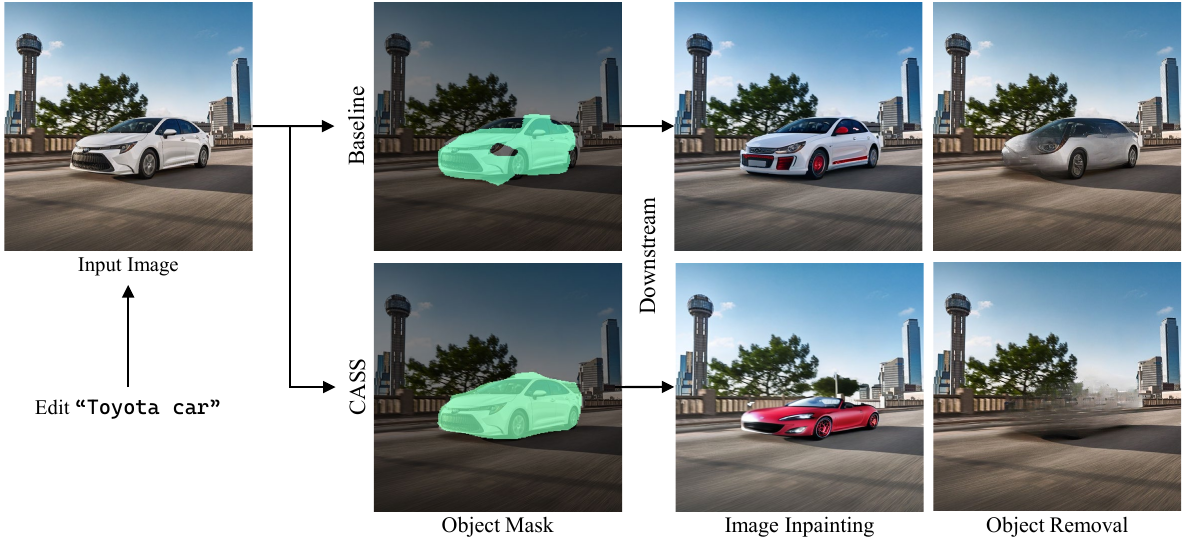}

   \caption{Visualization of image inpainting and object removal using our predicted mask. For image inpainting, we use ``\texttt{red sports car with red wheels}" as an input prompt. Note that the mask refinement step is excluded when segmenting the object mask.
   }
   \label{fig:inpainting}
\end{figure*}
}

While our {\ourmethod} can naturally be used for semantic segmentation directly based on user-provided prompts in real-world scenarios (see Sec.~\ref{sec:supp_realworld} for diverse examples), its ability to be object-level context-aware allows it to group object components effectively. 
As a result, {\ourmethod} produces \textit{clean object masks} that can be effectively utilized in downstream tasks (e.g., image inpainting and object removal).  
Fig.~\ref{fig:inpainting} visualizes the application of our {\ourmethod} in various downstream tasks. 
When provided with the object prompt \texttt{Toyota car}, our model generates a precise object mask, enabling tasks such as image inpainting and object removal. 
For image inpainting, we utilize Stable Diffusion XL (SDXL), while for object removal, we employ LaMa~\cite{suvorov2022resolution}. The baseline model used for comparison is NACLIP~\cite{hajimiri2025naclip}.  
Comparing the results to the baseline, it is evident that the baseline model struggles to correctly mask all the object components, such as the \textit{rear wheels} and \textit{headlights}. 
For example, the baseline fails to mask the rear wheel of the car, which limits the inpainting model from accurately rendering the red wheel in the edited image. Similarly, incomplete masking of headlights results in unedited regions (e.g., headlight) during the inpainting process.
This limitation also affects object removal tasks.
The baseline's failure to mask the entire object (i.e., \texttt{Toyota car}) limits object-level removal and instead focuses on specific internal components, such as side mirrors or door handles.
In contrast, our method successfully captures all relevant object components, producing a complete and accurate object mask that enables the removal of the object as a whole.
These results demonstrate the importance of generating object-level context-aware masks, as they enable downstream models to perform tasks such as image inpainting and object removal with significantly improved precision and quality.

\subsection{Open-Vocabulary Semantic Segmentation in the Wild}
\begin{figure*}[ht]
  \centering
   \includegraphics[width=0.87\linewidth]{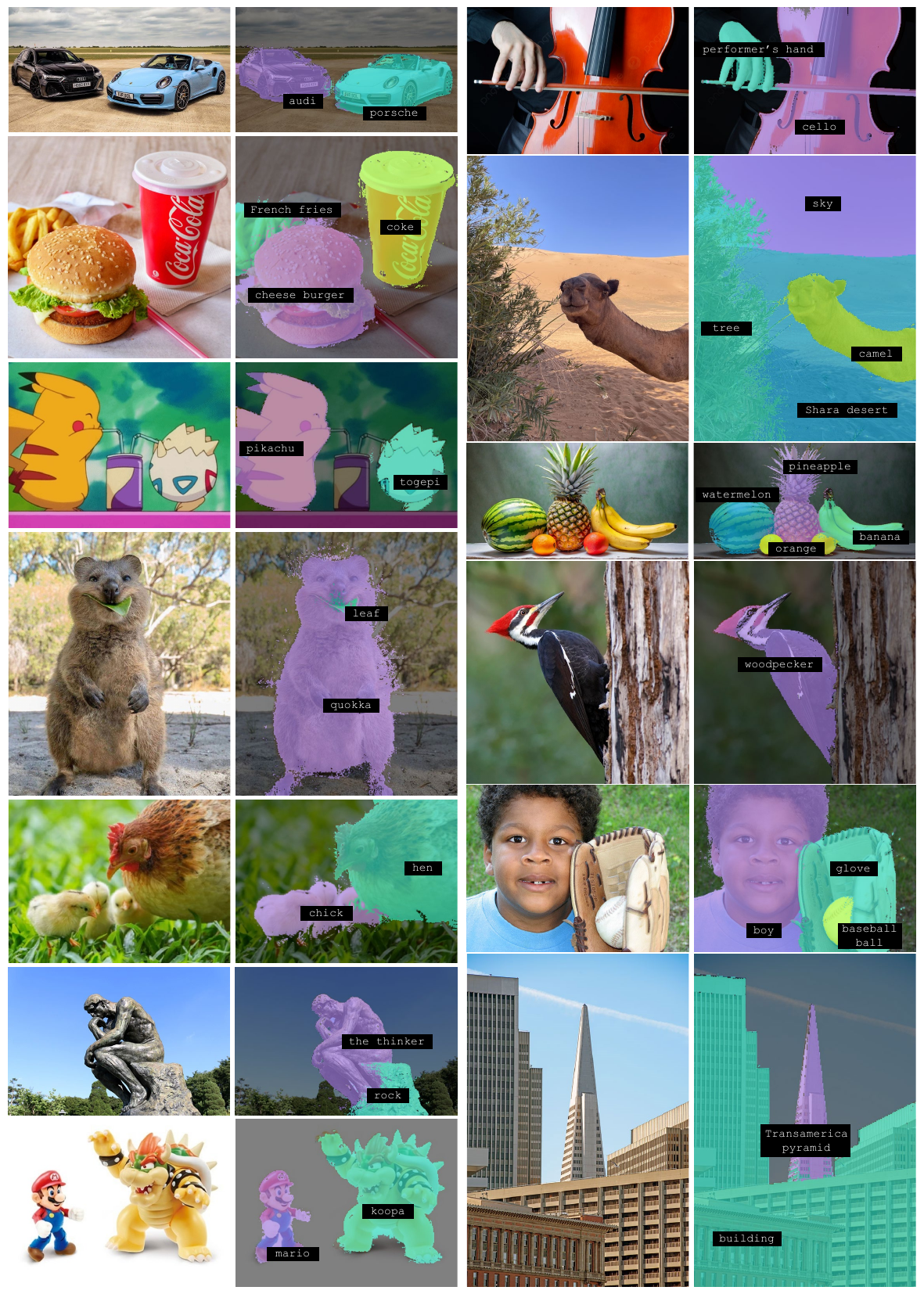}
   \caption{Real-world open-vocabulary semantic segmentation results of our model after applying mask refinement~\cite{araslanov2020single}.}
   \label{fig:real}
\end{figure*}

Fig.~\ref{fig:real} illustrates the application of {\ourmethod} on real-world images, demonstrating its ability to adapt to diverse and uncontrolled settings. 
This showcases the method's robustness in handling the complexities and variations typically encountered in real-world scenarios, without requiring additional fine-tuning or preprocessing.

\label{sec:supp_realworld}

\section{Discussion}
\subsection{Limitations}

\begin{figure}[ht]
  \centering
   \includegraphics[width=\linewidth]{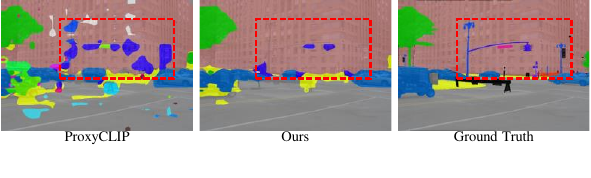}
   \vspace{-25pt}
   \caption{Visualization of limitation of {\ourmethod}.
   }
   \label{fig:limitation}
\end{figure}

Our {\ourmethod} introduces object-level contextual knowledge into training-free OVSS, enabling the effective grouping of object components into coherent semantic entities. By combining Energy-based Low-rank Approximation with spectral-based graph matching, our approach narrows the gap between pixel-level predictions and object-level understanding, contributing a significant advancement towards the primary objectives of OVSS.

However, {\ourmethod} has limitations, particularly in computational efficiency. While Energy-based Low-rank Approximation reduces costs, the eigendecomposition step remains computationally demanding, especially for high-resolution images. Additionally, the Hungarian matching algorithm, which runs on the CPU, introduces further latency due to its lack of GPU support.
These inefficiencies hinder real-time applicability. For instance, {\ourmethod} processes at 5.6 FPS on an RTX A6000 GPU, which is insufficient for real-time segmentation. Optimizing computational efficiency is essential to expanding its practical use in time-sensitive tasks.

Moreover, as shown in Fig.~\ref{fig:limitation}, our emphasis on object-level context leads to stronger performance on large objects (e.g., building: 20.0~\cite{lan2024proxyclip} to 34.7 mIoU) but can underperform on smaller objects, ultimately diminishing the overall mIoU improvement due to the averaging across all classes. As a result, {\ourmethod} yields substantial improvements on datasets with predominantly large and distinct object classes (e.g., VOC), while showing relatively modest gains on datasets rich in small object categories (e.g., ADE, COCO). Nonetheless, {\ourmethod} still secures a 7.25\% relative improvement over the existing SoTA~\cite{lan2024proxyclip} on average across 8 datasets, underscoring its meaningful contribution to advancing training-free OVSS.

\subsection{Future Works}
Our proposed {\ourmethod} is designed to perform semantic segmentation on individual images based on user-provided arbitrary prompts, achieving competitive performance in grouping object components into coherent entities. While effective for static scenarios, the current approach does not account for temporal consistency, limiting its applicability to video sequences and dynamic environments. To extend its use to real-time or sequential analysis tasks, future work will focus on incorporating temporal information to ensure coherence across frames. This enhancement will enable the method to handle video streams more effectively, supporting applications such as object tracking and dynamic scene understanding.  

As mentioned in the limitations, the high computational time remains a challenge that must be addressed as part of our future work. Specifically, optimizing the eigendecomposition process and the Hungarian matching algorithm used for graph matching is crucial. These improvements will play a key role in enhancing the efficiency of our method and making real-time inference feasible for practical applications.

\clearpage
{
\begin{figure*}[ht]
  \centering
   \includegraphics[width=0.9\linewidth]{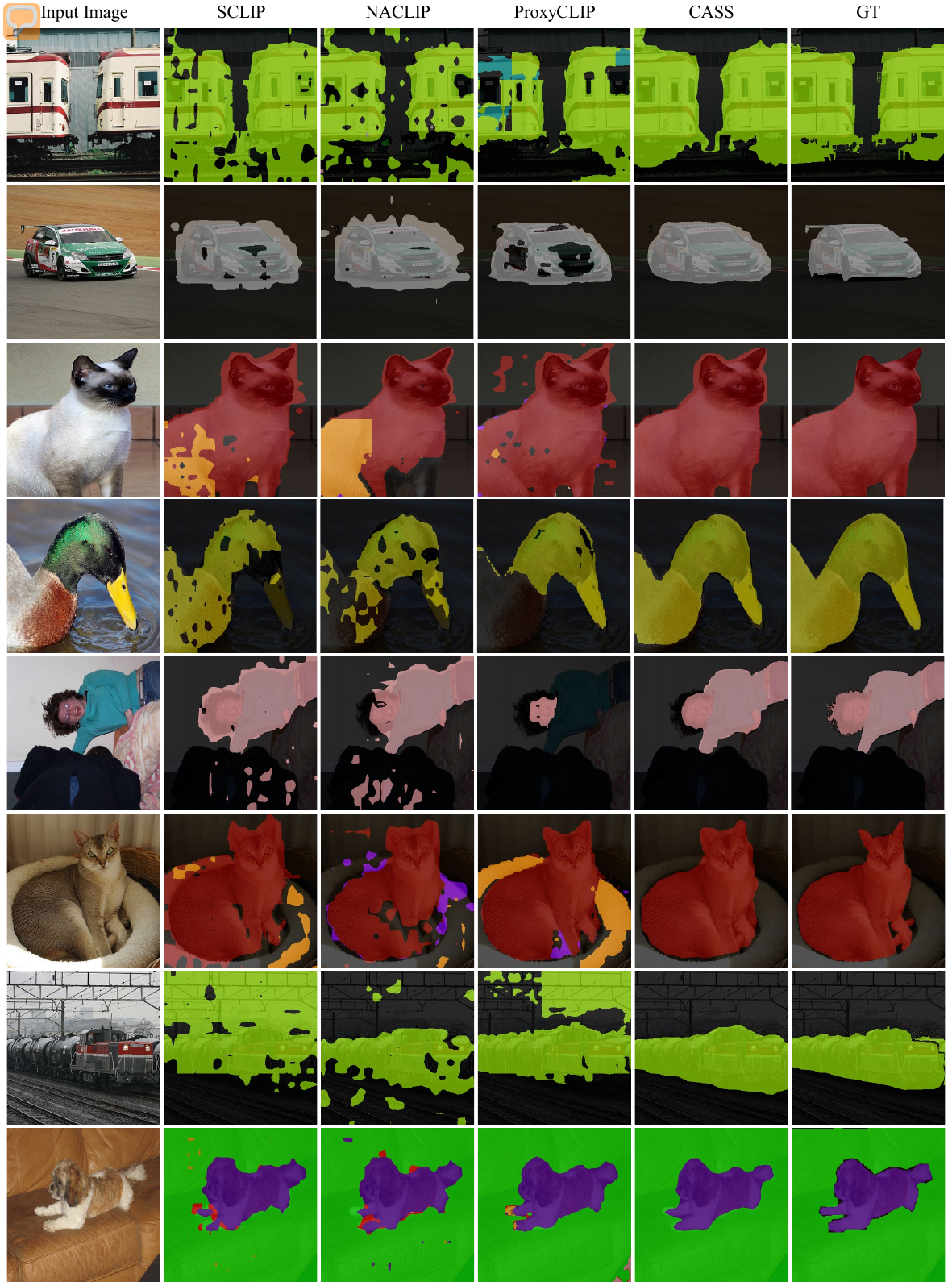}
   \caption{Additional qualitative comparison between recent state-of-the-art methods SCLIP~\cite{wang2023sclip}, NACLIP~\cite{hajimiri2025naclip}, ProxyCLIP~\cite{lan2024proxyclip} and Our {\ourmethod} using PASCAL VOC dataset~\cite{pascal-voc-2012}.
   }
   \label{fig:supp_voc}
\end{figure*}
}

{
\begin{figure*}[ht]
  \centering
   \includegraphics[width=0.9\linewidth]{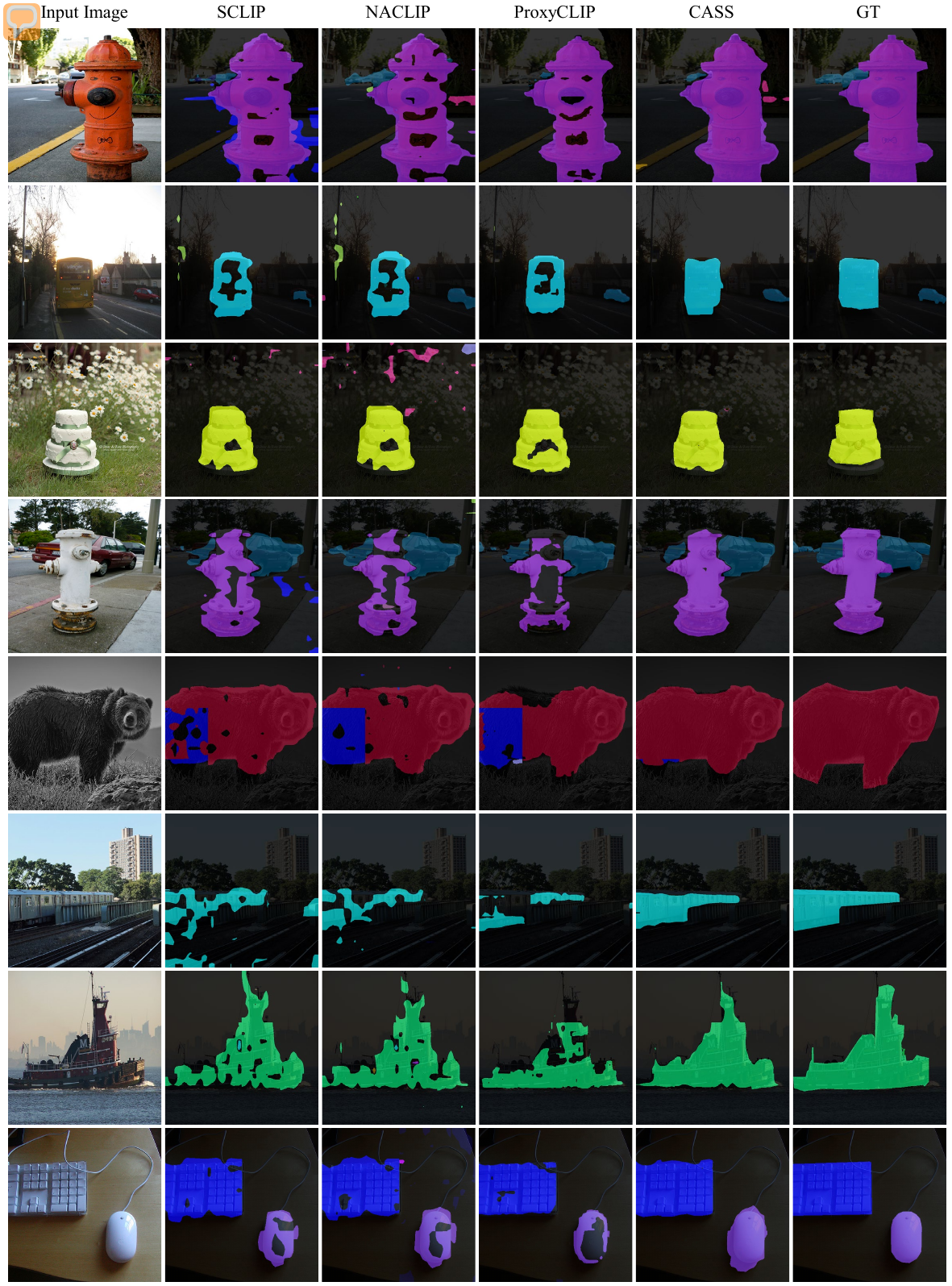}
   \caption{Additional qualitative comparison between recent state-of-the-art methods SCLIP~\cite{wang2023sclip}, NACLIP~\cite{hajimiri2025naclip}, ProxyCLIP~\cite{lan2024proxyclip} and Our {\ourmethod} using COCO dataset~\cite{caesar2018coco}.
   }
   \label{fig:supp_coco}
\end{figure*}
}

{
\begin{figure*}[ht]
  \centering
   \includegraphics[width=0.9\linewidth]{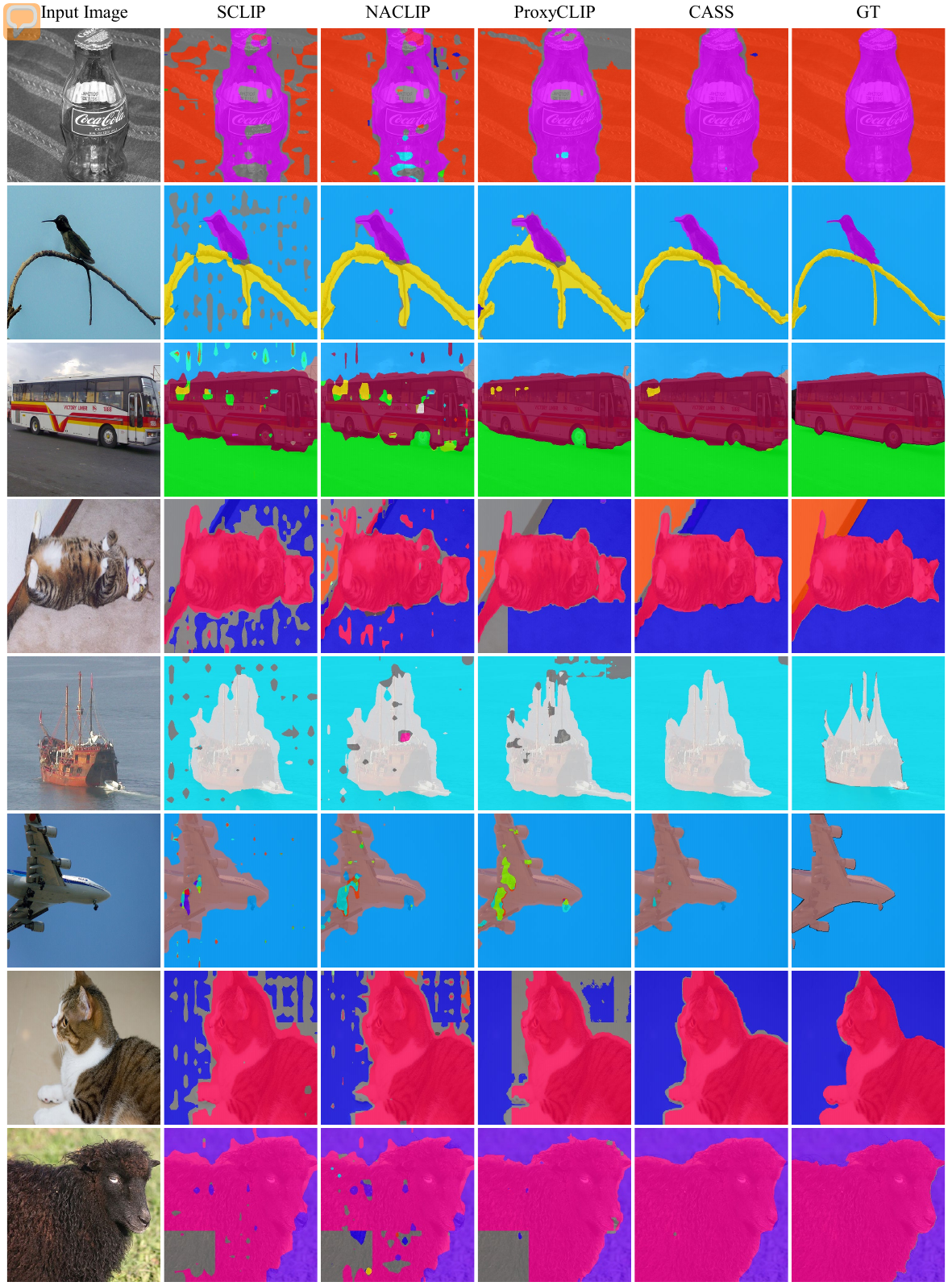}
   \caption{Additional quantitative comparison between recent state-of-the-art methods SCLIP~\cite{wang2023sclip}, NACLIP~\cite{hajimiri2025naclip}, ProxyCLIP~\cite{lan2024proxyclip} and Our {\ourmethod} using PASCAL Context~\cite{mottaghi2014role}.
   }
   \label{fig:supp_pascal}
\end{figure*}
}

{
\begin{figure*}[ht]
  \centering
   \includegraphics[width=0.9\linewidth]{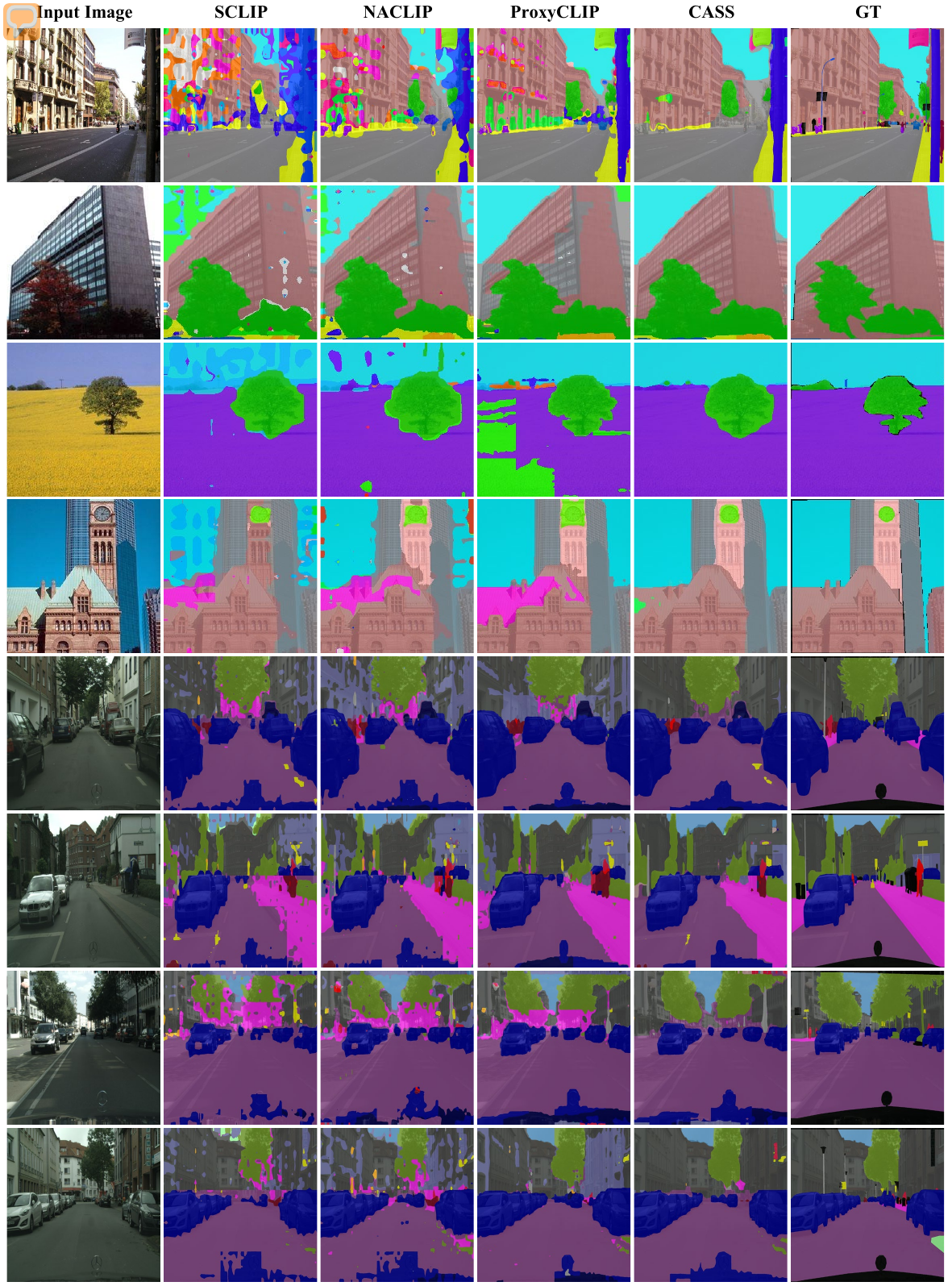}
   \caption{Additional qualitative comparison between recent state-of-the-art methods SCLIP~\cite{wang2023sclip}, NACLIP~\cite{hajimiri2025naclip}, ProxyCLIP~\cite{lan2024proxyclip} and Our {\ourmethod} using ADE20K~\cite{zhou2019semantic} (top) and Citscapes~\cite{cordts2016cityscapes} (bottom).
   }
   \label{fig:supp_adecity}
\end{figure*}
}

\end{document}